%% file: neurips_2026.tex
\documentclass{article}

 \usepackage[preprint]{neurips_2026}

\usepackage[utf8]{inputenc} 
\usepackage[T1]{fontenc}    
\usepackage{hyperref}       
\usepackage{url}            
\usepackage{booktabs}       
\usepackage{amsfonts}       
\usepackage{nicefrac}       
\usepackage{microtype}      
\usepackage{xcolor}         
\usepackage{xspace}    
\usepackage{amsmath}
\usepackage{graphicx} 
\usepackage{multirow}

\usepackage{xspace}
\usepackage{float}

\usepackage{tabularx}
\usepackage{tcolorbox} 
\tcbuselibrary{skins,raster}
\usepackage{pifont}
\definecolor{regdark}{HTML}{225C2E}
\definecolor{regsoft}{HTML}{EEF6E4}
\definecolor{emomag}{HTML}{B8327C}
\definecolor{emodark}{HTML}{3F1052}
\definecolor{emosoft}{HTML}{FBE7EF}
\definecolor{goldcol}{HTML}{1F6E7C}
\definecolor{goodgreen}{HTML}{2F8F3F}
\definecolor{nogray}{HTML}{888888}

\newcommand{\methodname}{\textsc{Emo}\xspace}
\newcommand{\selectiveeval}{selective expert use}
\newcommand{\Selectiveeval}{Selective expert use}
\newcommand{\SELECTIVEEVAL}{Selective Expert Use}

\let\oldparagraph\paragraph
\renewcommand{\paragraph}[1]{\vspace{-.5em}
\oldparagraph{#1}
}

\newif\ifcomments
\commentstrue
\ifcomments
    \newcommand{\rw}[1]{{\color{red}[ryan: #1]}}
    \newcommand{\sewon}[1]{{\color{violet}[sewon: #1]}}
\else
    \newcommand{\rw}[1]{}
    \newcommand{\sewon}[1]{}
\fi

\usepackage{xcolor}

\title{\methodname: Pretraining Mixture of Experts\\for Emergent Modularity

}

\definecolor{olmoePink}{HTML}{f0539b}
\definecolor{darkblue}{HTML}{254fc9}

\newcommand{\bb}{{\color{darkblue}\boldsymbol{\alpha}}}
\newcommand{\pa}{{\color{olmoePink}\boldsymbol{\beta}}}

\newcommand{\huggingface}{\raisebox{-1.5pt}{\includegraphics[height=1.05em]{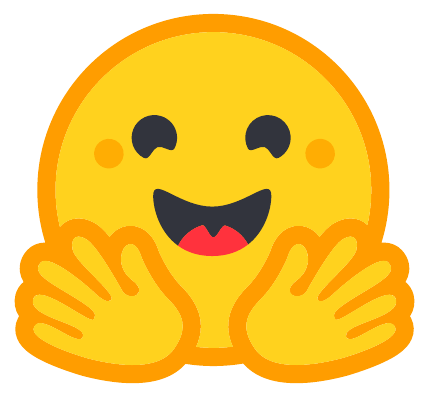}}\xspace}
\newcommand{\github}{\raisebox{-1.5pt}{\includegraphics[height=1.05em]{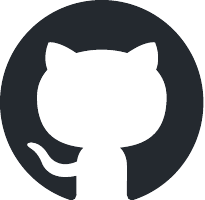}}\xspace}
\newcommand{\blog}{\raisebox{-1.5pt}{\includegraphics[height=1.05em]{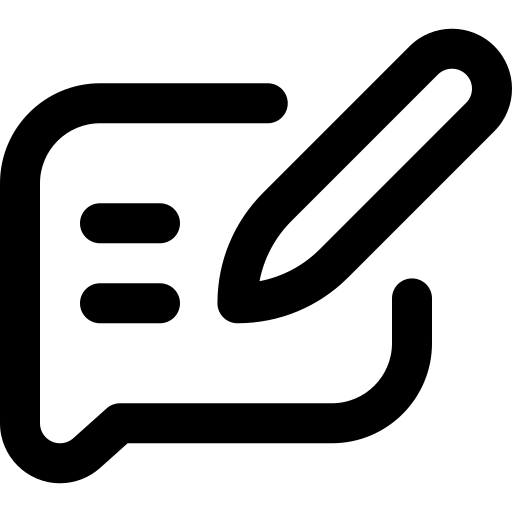}}\xspace}
\newcommand{\search}{\raisebox{-1.5pt}{\includegraphics[height=1.05em]{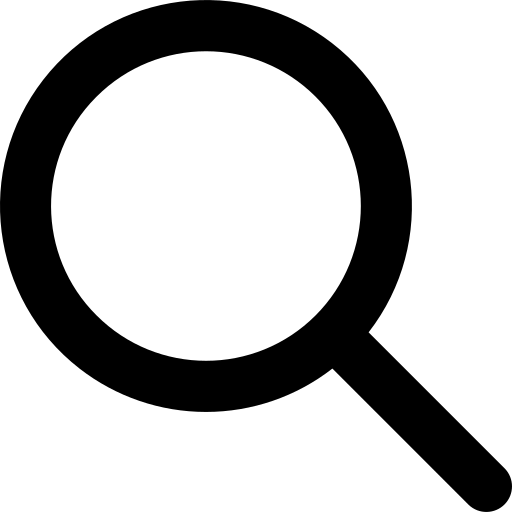}}\xspace}

\newcommand{\bw}

\author{%
  Ryan Wang$\hspace{.1em}^{\bb\pa}$ \hspace{.3em} Akshita Bhagia$\hspace{.1em}^{\pa}$\hspace{.3em} Sewon Min$\hspace{.1em}^{\bb\pa}$ \vspace{.5em} \\
  $\hspace{.1em}^{\bb}$University of California, Berkeley
  \quad
  $\hspace{.1em}^{\pa}$Allen Institute for AI
  \vspace{.5em} \\
  $\hspace{.1em}^{}$
  \texttt{ryanyxw@berkeley.edu} 
  \quad
  \texttt{akshitab@allenai.org} 
  \quad
  \texttt{sewonm@berkeley.edu} \\
}

\begin{document}

\maketitle

\vspace{-1.5em}
\begin{center}\small
    \begin{tabular}{rll}
        \huggingface & \textbf{Model} & \href{https://huggingface.co/collections/allenai/emo}{\path{hf.co/collections/allenai/emo}}\\[0.2em]
        \github & \textbf{Code} & \href{https://github.com/allenai/EMO}{\path{github.com/allenai/EMO}}\\[0.2em]
        \blog & \textbf{Blog} & \href{https://allenai.org/blog/emo}{\path{allenai.org/blog/emo}}\\ [0.2em]
        \search & \textbf{Visualization} & \href{https://emovisualization.netlify.app}{\path{https://emovisualization.netlify.app}}\\
    \end{tabular}
\end{center}
\vspace{1em}

\input{sections/00_abstract}

\section{Introduction}\label{sec:intro}\input{sections/01_intro}

\section{Related Work}\label{sec:relatedworks}

\input{sections/01_relatedwork}

\section{Modular Mixture of Experts (\methodname)}\label{sec:method}\input{sections/02_method}

\section{Experimental Setup}\label{sec:setup}\input{sections/02_setup}

\section{Results and Analysis}\label{sec:results}\input{sections/03_results}

\section{Future Directions}\label{sec:discussion}\input{sections/07_discussion}

\section{Conclusion}\label{sec:conclusion}\input{sections/conclusion}

\section*{Acknowledgement}

We thank Prasann Singhal, Gustavo Lucas Carvalho, Weijia Shi, Jagdeep Bhatia, Colin Raffel, Berkeley AI Research members, the Sky computing lab members, and Ai2 members for valuable discussion and feedback.

This research was supported in part by ONR (N00014-26-1-2233), the NVIDIA Academic Grant Program, and gifts from Ai2 and Apple.
Ryan Wang was supported by the National Science Foundation Graduate Research Fellowship Program. 

\setcitestyle{numbers,square}
\bibliographystyle{unsrt}
\bibliography{main.bib}


\newpage
\appendix\label{sec:appendix}\input{sections/06_appendix}


\newpage

\end{document}

%% file: sections/00_abstract.tex
\begin{abstract}

Large language models are typically deployed as monolithic systems, requiring the full model even when applications need only a narrow subset of capabilities, e.g., code, math, or domain-specific knowledge. Mixture-of-Experts (MoEs) seemingly offer a potential alternative by activating only a subset of experts per input, but in practice, restricting inference to a subset of experts for a given domain leads to severe performance degradation.
This limits their practicality in memory-constrained settings, especially as models grow larger and sparser. 
We introduce \textbf{\methodname}, an MoE designed for modularity—the independent use and composition of expert subsets—without requiring human-defined priors. Our key idea is to encourage tokens from similar domains to rely on similar experts. Since tokens within a document often share a domain, \methodname restricts them to select experts from a shared pool, while allowing different documents to use different pools. This simple constraint enables coherent expert groupings to emerge during pretraining using document boundaries alone.
We pretrain a 1B-active, 14B-total \methodname\ on 1T tokens. As a full model, it matches standard MoE performance. Crucially, it enables \selectiveeval: retaining only 25\% (12.5\%) of experts incurs just a 1\% (3\%) absolute drop, whereas standard MoEs break under the same setting. We further find that expert subsets in \methodname specialize at semantic levels (e.g., domains such as math or code), in contrast to the low-level syntactic specialization observed in standard MoEs. Altogether, our results demonstrate a path toward modular, memory-efficient deployment of large, sparse models and open new opportunities for composable architectures.

\end{abstract}

%% file: sections/01_intro.tex
Large language models (LLMs) are typically trained and deployed as monolithic systems: a single model is pretrained, finetuned, and served as one unified entity \cite{olmo2026olmo3, deepseekai2025deepseekv3technicalreport, yang2025qwen3technicalreport}. While effective, this paradigm becomes increasingly restrictive as models scale. In many deployment settings, applications require only a narrow subset of capabilities---such as code generation, mathematical reasoning, or domain-specific knowledge---but must still serve the full model, incurring unnecessary computational cost and memory use. Moreover, the monolithic design prevents isolating, updating, or improving specific capabilities without retraining and redeploying the entire system.

Mixture-of-Experts (MoE) models appear to offer a natural path toward relaxing this constraint, as they consist of many small FFNs ({\em experts}), of which only a small subset is activated for each input token \cite{deepseekai2025deepseekv3technicalreport, deepseekai2024deepseekv2strongeconomicalefficient}. However, existing MoEs still require the full model for any task: tokens within the same input activate different experts, causing most or all experts to be used over the course of a task. As we show, this behavior---partially driven by experts specializing in low-level lexical patterns (e.g., prepositions, punctuation)---prevents subsets of the model from being usable independently, limiting the deployability of MoEs in memory-constrained settings, an issue that becomes increasingly important as models grow larger and sparser \cite{dai2024deepseekmoeultimateexpertspecialization, deepseekai2025deepseekv3technicalreport, yang2025qwen3technicalreport}.

\input{figures/eval_flowchart}

We instead seek to train MoE models in which experts organize into coherent groups that can be selectively used and composed. Concretely, we train an MoE model to be {\em modular}, i.e., to support (1) the independent use of expert subsets and (2) their composition into a strong general-purpose model. 
Achieving this in practice, however, is challenging. Prior work has explored partitioning training data into predefined domains (e.g., math, coding) and training separate experts~\cite{sukhbaatar2024branchtrainmixmixingexpertllms, shi2025flexolmoopenlanguagemodels}, but this is too restricted for model's learning and limits the model's overall performance.

In this work, we propose to train MoE models in which modular structure emerges directly from the data, without relying on human-defined prior, and introduce \textbf{\methodname}, an MoE that follows this approach. Our key intuition is that tokens from the similar domains should activate similar subsets of experts.
Assuming that tokens within a document tend to share a domain, we enforce this structure by restricting all tokens in a document to select their active experts from a shared pool. For example, in an MoE with 128 total and 8 active experts, all tokens from a document select their active subset from a shared pool of 32 experts. Different documents may use different expert pools, allowing the model to learn recurring expert subsets across the training corpus. Importantly, \methodname does not require predefined task or domain labels: expert subsets emerge in a self-supervised way, using document boundaries as the only grouping signal.

We train a 1B-active, 14B-total parameter \methodname\ model on 1 trillion tokens. As a full model, \methodname\ matches the overall performance of a standard MoE. More importantly, however, it enables effective composition of expert subsets, which standard MoEs fail to support. Across domain-specific subsets of MMLU and MMLU-Pro (e.g., math, physics, biology, social sciences), identifying and deploying only the most relevant experts largely preserve performance, e.g., 1\% absolute performance drop when retaining 25\% of experts, and 3\% when retaining 12.5\%. This is in contrast to standard MoEs that see severe degradation under the same constraint, e.g., 10\% and 15\% drops, respectively. These results show that \methodname\ makes MoEs significantly more practical and accessible: instead of loading the full model, one can serve only a small subset of experts relevant to a given task or domain (Figure~\ref{fig:eval_flowchart}), which has important implications for deployment in memory-constrained settings \cite{song2025blockffnendsideaccelerationfriendlymixtureofexperts, shen2026temporally, tairin2025emoe}.

We further analyze routing patterns and find that expert subsets specialize at higher-level semantic granularity, such as domains and topics (e.g., math, code), which is in contrast to experts in standard MoEs that specialize in lower-level syntactic patterns (e.g., prepositions, punctuation). This difference suggests that expert specialization in \methodname\ is qualitatively distinct and underlies its modularity.


Together, these results demonstrate that modularity can be built into large language models, opening a path for broader functionalities, such as targeted extension training or more interpretable and debuggable components to better regulate model behavior.
We release both \methodname and a matched baseline trained on the same data to support reproducibility and further study. 

\vspace{-.5em}

%% file: figures/eval_flowchart.tex
\begin{figure*}[t]
\centering
\includegraphics[width=1\linewidth]{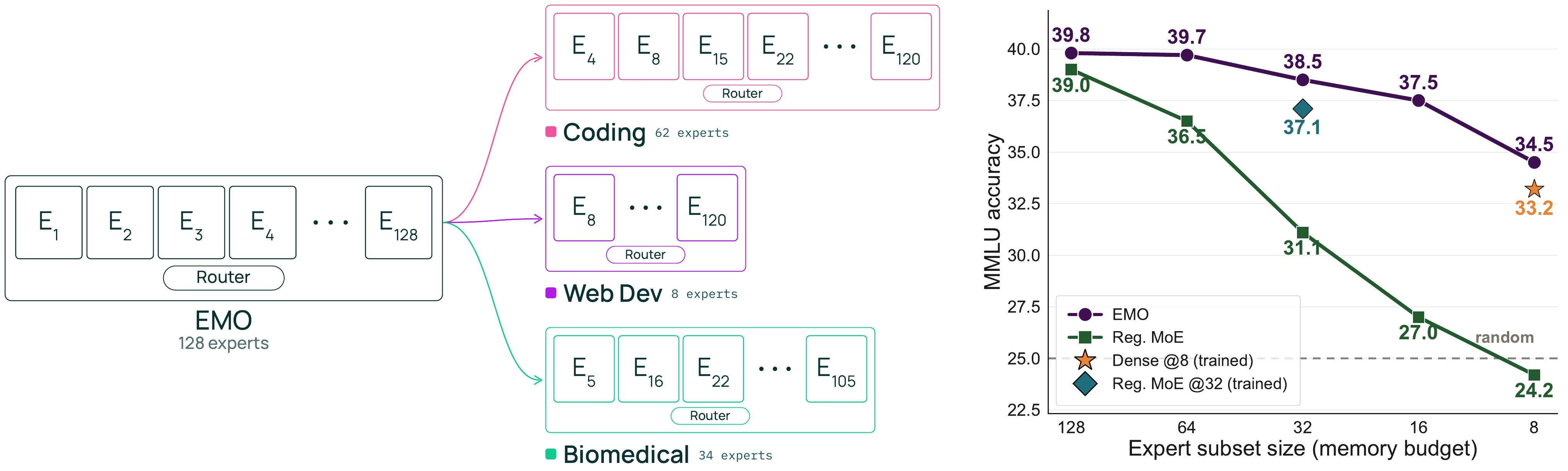}
\vspace{-.5em}
\caption{
\textbf{(Left)}
\methodname\ is an MoE trained with modularity as a first-class objective.
For a given domain (e.g., math, code, biomedical), users can select a small subset of experts of any size and retain near full-model performance. This turns a single model into a composable architecture, enabling flexible deployment with improved memory-accuracy tradeoffs for large, sparse MoEs.
\textbf{(Right)}
Averaged performance over 16 MMLU categories across different memory budgets. \methodname (purple) and Reg. MoE (green) are single models evaluated with expert subsets of different sizes. \methodname expert subsets push the Pareto frontier in memory-accuracy trade-off, outperforming standard MoEs and even fixed-budget models trained from scratch. 
}
\vspace{-.8em}
\label{fig:eval_flowchart}
\end{figure*}

%% file: sections/01_relatedwork.tex
\paragraph{Mixture-of-Experts as Scalable Architectures.}
Mixture-of-Experts (MoE) architectures introduce sparsity into Transformers by activating only a subset of experts per input, enabling efficient scaling to very large models \cite{shazeer2017outrageouslylargeneuralnetworks, lepikhin2020gshard, fedus2022switch}. Recent systems push this paradigm further by increasing both the number of experts and the degree of sparsity---for example, DeepSeek-V3 \cite{deepseekai2025deepseekv3technicalreport} employs hundreds of experts per layer while activating only a small subset per token---allowing models to reach scales of hundreds of billions of parameters.

As MoEs grow larger and sparser, memory bottlenecks become a central challenge: even inactive experts need to reside in VRAM at inference time. This has motivated a line of work such as memory-constrained scaling laws~\cite{li2026can}, memory-efficient serving~\cite{song2025blockffnendsideaccelerationfriendlymixtureofexperts, shen2026temporally, tairin2025emoe}, and expert pruning for a general purpose model that removes redundant experts \cite{lu2024expertsequalefficientexpert}.

This work introduces an MoE that enables selective use of expert subsets for a given downstream task. Among its benefits, this provides a new way to alleviate memory bottlenecks in large, sparse MoEs, complementary to prior approaches.

\paragraph{Specialization and Modularity of Existing MoEs.}
A growing body of work studies the extent to which specialization emerges in MoE models.
Prior work finds that specialization is often driven by surface-level patterns (e.g., token ID that is context-independent) or low-level lexical cues (e.g., prepositions, punctuations)~\cite{jiang2024mixtral, muennighoff2025olmoeopenmixtureofexpertslanguage}, while other works find that specialization is confined to only a tiny subset of experts~\cite{chaudhari2026moelensexpert}.
Other work suggests that apparent expert specialization may largely reflect geometric properties of the representation space that is difficult to interpret \cite{wang2026myth}.
In parallel, several works attempt to exploit these patterns for efficiency, for example by pruning experts for a given task~\cite{hu2025quantifyingexpertspecializationforeffectivepruning, dong2025domainspecificpruninglargemixtureofexperts, lu2024expertsequalefficientexpert, chen2022taskspecificexpertpruningsparse,huang2026discovering}.

In this work, we show that standard MoEs trained with conventional objectives do not support meaningful use of small expert subsets for downstream domains, and instead advocate for training an MoE with modularity as a first-class objective. When training accordingly, MoEs naturally support selective use of expert subsets, and this behavior is robust across different subset selection methods.

\paragraph{Training MoEs with Structured or Specialized Experts.}
Prior work has explored training MoEs with more structured or specialized experts. One line of work promotes interpretability or diversity across experts, primarily to reduce redundancy~\cite{yang2025mixtureexpertsintrinsicallyinterpretable, park2025monetmixturemonosemanticexperts,hu2026improvingmoeperformancewithplugandplay, guo2026advancingexpertspecializationbetter}, but such approaches do not ensure that expert subsets are usable in isolation. Another line of work explicitly partitions training data into predefined domains (e.g., math, biomedical), train separate experts, and merge them into a single MoE~\cite{shi2025flexolmoopenlanguagemodels, sukhbaatar2024branchtrainmixmixingexpertllms, li2022branchtrainmergeembarrassinglyparalleltraining}. While this enables standalone use of expert subsets, it relies on fixed, human-defined priors, which restricts flexibility and limits overall model performance. In contrast, we train an MoE end-to-end with modularity as a first-class objective, allowing expert structure to emerge without requiring predefined domains or human priors.

The closest line of work is ModuleFormer \cite{shen2023moduleformermodularityemergesmixtureofexperts}, which shares our goal of training a modular MoE that supports standalone use of expert subsets. It introduces an objective that maximizes mutual information between tokens and experts. However, they evaluate only against dense models, without standard MoEs. We attempted to reproduce ModuleFormer and found that they do not perform better than standard MoEs, and degrades significantly when less than 40\% of experts are retained, which is consistent with their reported results. \methodname largely shares the motivation with ModuleFormer but proposed a more effective training objective that significantly outperforms standard MoEs and other parameter-matched and memory-matched baselines, showing minimal degradation even with an expert subset size of just 12.5\%.

%% file: sections/02_method.tex
\input{figures/modmoe_method}

The goal of \methodname\ is to pre-train an MoE with modularity as the first-class objective, i.e., (1) expert subsets should be usable in isolation for a particular downstream domain, and (2) their composition---the full model---remains a strong general-purpose model.

\paragraph{Naive Approach.} A straightforward approach to develop modularity is to enforce expert specialization in MoEs by routing tokens to experts based on predefined semantic domains (e.g., math, biology, code). Methods such as FlexOlmo \cite{shi2025flexolmoopenlanguagemodels} and BTX \cite{sukhbaatar2024branchtrainmixmixingexpertllms} instantiate this idea. However, this formulation requires domain labels across pretraining data, which can be ambigious, difficult to obtain, and injects human biases. Having fixed domains also restricts flexibility, making it difficult for the model to be applied to new domains during inference.  

\paragraph{\methodname's Approach.}
Instead, we induce modular structure without explicit domain labels (Figure~\ref{fig:modmoe_method}). Our key observation is that tokens within the same document usually come from the same domain. We therefore treat \emph{document boundaries} as a \emph{weak supervisory signal}: for each document, the router selects a shared expert pool, and all tokens in that document choose their active experts only from this pool. Different documents can use different pools, allowing modular expert subsets to emerge directly from the training data. 

In the rest of the section, we first describe the standard MoE architecture and objective (\S\ref{subsec:moe_obj}), then describe \methodname's training objective (\S\ref{subsec:ours_obj}).

\subsection{Preliminary: Mixture of Experts Architecture}\label{subsec:moe_obj}

Mixture-of-Experts (MoE) models are decoder-only Transformer language models \citep{vaswani2023attentionneed} in which the feedforward sublayer is replaced by a sparse mixture of expert networks. Let the model contain $n$ total experts, consisting of $n_r$ routed experts and $n_s$ shared experts ($n = n_r + n_s$). Routed experts are selected dynamically on a per-token basis, while shared experts are always active.

Given the hidden state $x_t$ at token position $t$, a router produces logits over the routed experts,
\[
r(x_t) \in \mathbb{R}^{n_r}, \qquad
p_t = \operatorname{softmax}(r(x_t)).
\]
Let $\mathcal{K}_t = \operatorname{Top}\text{-}K(p_t, k) \subseteq \{1, \dots, n_r\}$ denote the indices of the top-$k$ routed experts selected for token $t$. The MoE feedforward output is then
\[
\operatorname{FFN}_{\mathrm{out}}(x_t)
=
\sum_{i \in \mathcal{K}_t} (p_t)_i \, E_i(x_t)
\;+\;
\sum_{j=1}^{n_s} E^{(s)}_j(x_t),
\]
where $E_i$ denotes the $i$-th routed expert and $E^{(s)}_j$ denotes the $j$-th shared expert.

The resulting $\operatorname{FFN}_{\mathrm{out}}(x_t)$ is used throughout the forward pass of the model to compute token probabilities. We train the model using the standard autoregressive language modeling objective:
\[
\mathcal{L}_{\mathrm{CE}} = -\sum_{t=1}^{T} \log P(x_t \mid x_{<t}),
\]
where the conditional probabilities $P(x_t \mid x_{<t})$ are computed using the MoE layer defined above.

In addition to the cross entropy, MoE training includes auxiliary losses such as the load balancing loss $\mathcal{L}_{\mathrm{LB}}$ to encourage uniform expert utilization: \[
\mathcal{L}_{\mathrm{LB}} = n_r \sum_{i=1}^{n_r} \bar{f}_i \cdot \bar{P}_i,
\] where $\bar{f}_i$ is the fraction of tokens routed to expert $i$ and $\bar{P}_i$ is the average routing probability of expert $i$ across all tokens. 
%
The full objective is\[
\mathcal{L}
=
\mathcal{L}_{\mathrm{CE}}
+
\alpha \mathcal{L}_{\mathrm{LB}}
+
\beta \mathcal{L}_{\mathrm{RZ}},
\]
where $\mathcal{L}_{\mathrm{RZ}}$ regularizes router logits, and $\alpha$ and $\beta$ control auxiliary loss weights. 

\subsection{\methodname: An Objective to Induce Modularity} \label{subsec:ours_obj}

The goal of \methodname\ is to induce modularity by leveraging \emph{document boundaries} as a weak supervisory signal. \methodname\ achieves this by selecting a \emph{document expert pool} for each document and constrains all tokens in the document to route within this pool during training (Figure~\ref{fig:modmoe_method}). 

\paragraph{Formulation.}
Recall that $p_t = \operatorname{softmax}(r(x_t)) \in \mathbb{R}^{n_r}$ denotes the routing distribution for token $t$.
We define the document expert pool $\mathcal{D}$ based on the average routing distribution across tokens: 
\[
\mathcal{D} = \operatorname{Top}\text{-}K\!\left( \frac{1}{T} \sum_{t=1}^{T} p_t, \, d \right)
\subseteq \{1, \dots, n_r\}.
\]
Routing is then restricted to $\mathcal{D}$ via a masked and renormalized distribution
\[
\hat{p}_t(i)
=
\begin{cases}
\dfrac{p_t(i)}{\sum_{j \in \mathcal{D}} p_t(j)} & \text{if } i \in \mathcal{D}, \\[8pt]
0 & \text{otherwise}.
\end{cases}
\]
The routed experts are then
\[
\mathcal{R}_t = \operatorname{Top}\text{-}K(\hat{p}_t, k).
\]

The resulting feedforward output is
\[
\operatorname{FFN}_{\mathrm{out}}(x_t)
=
\sum_{i \in \mathcal{R}_t} (\hat{p}_t)_i \, E_i(x_t)
\;+\;
\sum_{j=1}^{n_s} E^{(s)}_j(x_t),
\]
where $E_i$ denotes routed experts and $E^{(s)}_j$ denotes shared experts.

The hyperparameter $d$ controls subset granularity: smaller $d$ enforces highly specialized expert subsets with limited expressivity (e.g., $d=k$ forces all tokens in a document to use the same experts), while larger $d$ increases flexibility at the cost of weaker modular structure (e.g., $d=n_r$ recovers the standard MoE).

\subsection{Key Technical Considerations}
\label{subsec:technical_considerations}

Several technical choices were important for effective training
of \methodname (see \S\ref{appendix:hyperparameters} for details). 

\paragraph{Consideration 1. Load Balancing.}

A central challenge is that load balancing and document-level routing appear to impose opposing pressures. This conflict arises under standard micro-batch load balancing, where the load-balancing loss is computed over only a few documents. While this local implementation reduces cross-device communication and simplifies distributed training, it also encourages tokens from the same document to spread across many experts, directly opposing the shared-pool constraint and causing unstable training.

We address this by adopting global load balancing \cite{qiu2025demonsdetailimplementingload}, aggregating routing statistics across data-parallel groups. Applied over a larger and more diverse set of documents, load balancing encourages uniform utilization of experts \textit{across} documents, while our routing constraint enforces expert consistency \textit{within} each document, making the two objectives largely complementary. Empirically, this is important for stable training: see Figure~\ref{fig:global_vs_local_lb} in \S\ref{appendix:hyperparameters}.

\paragraph{Consideration 2. Choosing Expert Pool Size.}
Fixing a single expert pool size $d$ works well during training but limits inference-time flexibility. The model "overfits" only to expert sets of size $d$ and performs poorly when deployed as expert subsets that isn't of size $d$.

To enable the model to support expert subsets of all sizes, we treat $d$ as a random variable and sample it independently for each document during pretraining:
$$
d \sim \mathcal{U}\{k, \dots, n_r\}.
$$
where $k$ is the number of active experts per token and $n_r$ is the total number of routable experts. This exposes the model to a range of expert pool sizes during training, enabling it to support expert subsets of varying capacities for \selectiveeval.

%% file: figures/modmoe_method.tex
\begin{figure*}[t]
\centering
\includegraphics[width=1\linewidth]{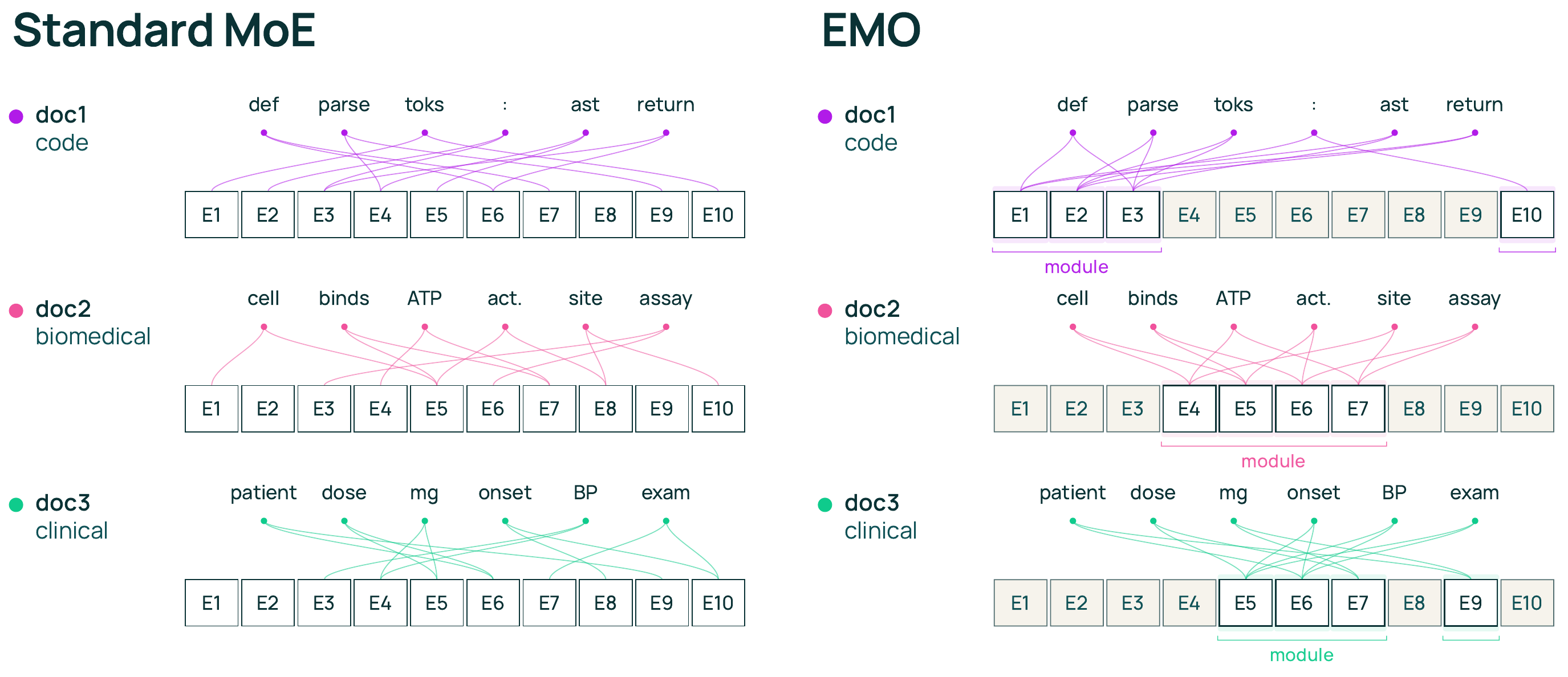}
\vspace{-1.5em}
\caption{
Comparison of training of a standard MoE and \methodname\ ($k=2, n=10$, shared experts omitted for simplicity).
\textbf{(Left)} In a standard MoE, each token independently selects its top-$k$ experts. 
\textbf{(Right)} In \methodname, the router first selects a subset of experts for each document, and all tokens are constrained to route within this subset. This enforces consistent expert usage across the document, encouraging groups of experts to form domain specialization.
}
\vspace{-.5em}
\label{fig:modmoe_method}
\end{figure*}

%% file: sections/02_setup.tex
\subsection{Architecture \& Training Details} \label{sec:model_details}

We consider an MoE with 1B active and 14B total parameters, consisting of $n=128$ experts ($n_r=127$ routed, $n_s=1$ shared), with $k=8$ experts activated per token.
The baseline MoE and \methodname\ share the same architecture; they differ only in their training objectives, as described in \S\ref{subsec:ours_obj}.

We train both the baseline MoE and \methodname\ from scratch on 1 trillion tokens from the OLMoE pretraining corpus \citep{muennighoff2025olmoeopenmixtureofexpertslanguage}, followed by an additional 50B-token linear annealing phase.
For ablations, we additionally train models on 130B tokens and include comparison to dense baselines and smaller MoEs.

Our architecture largely follows that of OLMoE~\citep{muennighoff2025olmoeopenmixtureofexpertslanguage}, with several key improvements: (1) adding a shared expert, (2) using pre-norm instead of post-norm, and (3) removing QK-norm; see \S\ref{appendix:hyperparameters} for details on the improvements introduced by these changes.
These modifications make our baseline MoEs significantly more competitive: as shown in \S\ref{subsec:full_model_eval}, our baseline MoE trained on 1T tokens consistently outperforms OLMoE trained on 5T tokens despite being trained on the same data.


\subsection{Evaluation}\label{subsec:eval}

We evaluate our models under two settings: (1) \textbf{full-model evaluation}, reflecting the standard use case in which a pretrained model is deployed for a broad set of tasks, and (2) \textbf{\selectiveeval}, where only a task-specific subset of experts is activated for a particular task or domain. Additional details on evaluation tasks and settings are provided in \S\ref{appendix:evaluation_detail}.

\paragraph{Full-model Evaluation.}
We first evaluate the full model under zero-shot settings. 
We report results on five evaluation suites:
(1) \textbf{\textsc{MC9}}, an average over nine multiple-choice benchmarks including ARC-Easy~\citep{allenai:arc}, ARC-Challenge~\citep{allenai:arc}, BoolQ~\citep{clark2019boolqexploringsurprisingdifficulty}, CSQA~\citep{talmor2019commonsenseqa}, HellaSwag~\citep{zellers2019hellaswagmachinereallyfinish}, OpenBookQA~\citep{OpenBookQA2018}, PIQA~\citep{bisk2019piqareasoningphysicalcommonsense}, SocialIQa~\citep{sap2019socialiqacommonsensereasoningsocial}, and WinoGrande~\citep{sakaguchi2019winogrande};
(2) \textbf{Gen5}, an average over five generative tasks including CoQA~\citep{reddy2019coqaconversationalquestionanswering}, SQuAD~\citep{rajpurkar2016squad100000questionsmachine}, Natural Questions~\citep{kwiatkowski-etal-2019-natural}, TriviaQA~\citep{joshi2017triviaqalargescaledistantly}, and DROP~\citep{dua2019dropreadingcomprehensionbenchmark};
(3) \textbf{MMLU}~\citep{hendryckstest2021}\footnote{Aggregated results exclude the ``other'' category; see \S\ref{appendix:evaluation_detail} and \ref{appendix:mmlu_other_ablation} for details.\label{fn:mmlu_other_detail}};
(4) \textbf{MMLU-Pro}~\citep{wang2024mmluprorobustchallengingmultitask}\footref{fn:mmlu_other_detail}; and
(5) \textbf{\textsc{GSM8K}}~\citep{cobbe2021trainingverifierssolvemath}.

\paragraph{\SELECTIVEEVAL.}
We next evaluate whether models can be deployed using only a subset of experts for each downstream domain (Figure~\ref{fig:eval_flowchart}).
We consider coarse-grained domain grouping of MMLU and MMLU-Pro, e.g., math, physics, health, philosophy, history, which contain 16\footref{fn:mmlu_other_detail} and 13\footref{fn:mmlu_other_detail} domains, respectively, as well as \textsc{GSM8K}.

For each domain, we assume access to a small validation set to identify relevant experts. In \S\ref{appendix:validation_set_ablations}, we show that this validation set can be extremely small: even a single few-shot example is sufficient to select an effective expert subset. We consider two selection methods: (1) a simple approach that aggregates routing probabilities across tokens and ranks experts by their average routing probability, and (2) Easy-EP \cite{dong2025domainspecificpruninglargemixtureofexperts}, a more computationally expensive, state-of-the-art expert selection method.
We then retain the top-$d$ experts in each layer and discard the rest, producing a domain-specific subset of experts that can be used as a standalone model.
We vary $d$ to measure how performance changes as fewer experts are retained. We report both zero-shot performance and performance after finetuning.
More evaluation details can be found in \S\ref{appendix:evaluation_detail}.

%% file: sections/03_results.tex
\subsection{Full-Model Evaluation}\label{subsec:full_model_eval}

\input{tables/full_model_evals}

Table~\ref{tab:full_model_evals} reports full-model performance for models trained on the same data with the same number of active parameters (1B).
First, our baseline MoE is competitive, outperforming \textsc{OLMoE} \cite{muennighoff2025olmoeopenmixtureofexpertslanguage} trained on 5T tokens despite using only 1T tokens. Nonetheless, \methodname\ matches the performance of this standard MoE.

The trend holds in the 130B-token setting: both our baseline MoE and \methodname\ significantly outperform a dense model with matched active parameters, demonstrating the benefits of sparsity. \methodname\ remains comparable to the standard MoE.

\subsection{\SELECTIVEEVAL}\label{subsec:selective_eval}

\input{figures/modular_use_1t}

We evaluate whether expert subsets in \methodname can retain full-model performance for a given domain (Figure~\ref{fig:modular_use_1t} for 1T tokens, Figure~\ref{fig:modular_use_130b} for 130B tokens). 

\paragraph{Standard MoEs (Green) Degrade Sharply.} Restricting to expert subsets leads to large performance drops---for instance, over 10\% when retaining 25\% of experts (128$\rightarrow$32), and below a dense model with matched active parameters (Figure~\ref{fig:modular_use_130b}).
This trend is consistent with and without fine-tuning. These results show that standard MoEs do not support modular use: even when only a narrow set of capabilities is required, restricting to expert subsets causes performance to break down.

\paragraph{\methodname\ (Purple) Enables Modular Use.}
In contrast, \methodname\ retains performance under subset deployment. Performance drops are minimal, e.g., about 1\% at 25\% expert retention and 3\% at 12.5\%, and the model continues to outperform dense baselines, even in an extreme case where only 6.2\% of experts are retained. This trend persists after fine-tuning, e.g., notably, on GSM8K, subsets with up to 12.5\% of experts perfectly recover full-model performance. These results indicate that \methodname\ supports modular use by identifying domain-relevant expert subsets, enabling significant memory savings by avoiding the need to load the full model.

Notably, we show in \S\ref{appendix:validation_set_ablations} that selecting relevant expert subsets is sample efficient, needing as few as five examples. We also provide examples of GSM8K generations in \S\ref{appendix:gsm8k_examples}, demonstrating qualitative differences in generation quality between expert subsets of \methodname and those for standard MoEs. 

\paragraph{Expert Subsets of \methodname\ Outperform Memory-matched Models Trained from Scratch.}

Figure~\ref{fig:eval_flowchart} (right) compares \methodname\ expert subsets against memory-matched models trained from scratch, including a standard MoE with 32 experts and a dense model. A full set of results can be found in Figure~\ref{fig:modular_use_130b} in \S\ref{appendix:selective_130B_ablation}. Despite using only a subset of the experts from a larger pretrained model, the 32-expert and 8-expert subsets of \methodname\ match or outperform these memory-matched baselines. These results suggest that expert subsets from a single \methodname\ model offer a stronger memory-accuracy trade-off than models trained from scratch under fixed memory budgets, forming a new Pareto frontier across memory regimes.

\paragraph{\methodname is Robust to Expert Selection Schemes.}
\input{figures/expert_selection_method}

In addition to router-based selection, we evaluate \textsc{Easy-EP} \cite{dong2025domainspecificpruninglargemixtureofexperts}, the state-of-the-art expert pruning method as an alternative selection strategy (Figure~\ref{fig:expert_selection_method}). As a sanity check, we also compare with random selection. 

For a standard MoE, Easy-EP consistently outperforms router-based selection when larger subsets are retained, although performance still degrades sharply as the subset size decreases, consistent with observations from the original paper~\cite{dong2025domainspecificpruninglargemixtureofexperts}. This suggests that even state-of-the-art selection methods cannot overcome the lack of localized domain-specific capabilities.

In contrast, \methodname\ achieves strong performance under both router-based and Easy-EP selection methods. Performance is largely insensitive to the choice of selection scheme, and remains robust even with small expert subsets.
This highlights that modularity must be learned during training, rather than recovered through post hoc expert selection.

In \S\ref{appendix:modular_anneal}, we also test whether modularity can emerge after pre-training by annealing a standard MoE with the document-level expert pool objective (\S\ref{subsec:ours_obj}); while training \methodname\ from scratch performs best, the annealed model shows signs of modularity, which we leave for future work to investigate. 

\subsection{
    Semantic Specialization Emerge in \methodname
}\label{subsec:semantic_specialization}

\input{figures/token_cluster_comparison}

\input{figures/domain_similarities}

We analyze how functional modularity emerges in \methodname\ by examining expert specialization. We find a qualitative shift in behavior: while standard MoEs specialize at the lexical level, \methodname\ induces specialization at the level of domains and topics.

To study this, we cluster pretraining tokens based on their routing behavior. Specifically, we sample the first 100 tokens from 12K documents and extract routing probabilities across experts. We project these representations using PCA (retaining 95\% variance), apply L2 normalization, and cluster them using spherical k-means with 32 clusters. 

\paragraph{\methodname Clusters Align with Semantic Themes.}

An interactive visualization is available at \href{https://emovisualization.netlify.app}{\texttt{emovisualization.netlify.app}}; 

Figure~\ref{fig:token_cluster_comparison} shows representative cluster descriptions, assigned by Claude Code. First, in a standard MoE, clusters correspond to low-level lexical categories, such as ``prepositions'', ``proper names'', ``copula verbs'', or ``definite articles'', consistent with observations from prior work \cite{jiang2024mixtral, muennighoff2025olmoeopenmixtureofexpertslanguage}. As a result, tokens within a single document are dispersed across many clusters. 

In contrast, \methodname\ produces clusters aligned with high-level semantics and domains, such as ``film, music, TV \& book reviews'', ``health, medical \& wellness'', ``news reporting'', and ``US. politics \& elections''. As a result, tokens within the same document are typically assigned to the same cluster, indicating consistent expert use.

This reveals two key insights. First, domain-level specialization emerges in \methodname, even with no explicit supervision. Second, this specialization directly enables modularity: tokens from the same domain share routing patterns, allowing computation to be localized to a small, coherent subset of experts. As a result, expert subsets can be used effectively for downstream domains.

\paragraph{\methodname\ Expert Activation Patterns across Domains are Distinct and Matches Human Intuition.}
We next ask whether domain-level expert activation patterns reflect human-interpretable domain similarity. We find that \methodname\ groups conceptually related domains while separating unrelated ones, a structure much less pronounced in standard MoEs.

To measure this, we use a random sample of 20 million documents from WebOrganizer \cite{wettig2025organizewebconstructingdomains}, which assigns each document to one of 24 human-labeled domains. For each domain, we construct a domain-level expert activation vector by first averaging router activations across tokens within each document, and then averaging these document-level vectors across all documents assigned to that domain. We measure similarity between domains using cosine similarity over the resulting domain-level expert activation vectors.

As shown in Figure~\ref{fig:domain_similarities}, \methodname\ produces domain-level similarity patterns that better match semantic relationships between human-labeled domains. Related domains exhibit higher expert-activation similarity, while unrelated domains are more clearly separated. In contrast, the standard MoE shows more diffuse similarities across domains, suggesting that its expert activations are less aligned with human-interpretable domain structure.

Across both models, early layers show limited domain structure, while later layers exhibit clearer alignment with human-labeled domains. This suggests that domain-level expert specialization emerges progressively in deeper layers, potentially motivating future work on inducing modularity in a layer-wise manner.

%% file: tables/full_model_evals.tex
\begin{table}[t]
\centering
\footnotesize
\begin{tabular}{l l r r r r r}
\toprule
 & \# train tokens & \textbf{MC9} & \textbf{Gen5} & \textbf{MMLU} & \textbf{MMLU Pro} & \textbf{GSM8K} \\
\midrule
    OLMoE$^\dagger$     & 5T & 63.5 & 57.6 & 42.8 & 18.7 & 13.7 \\
\midrule
    Reg. MoE            & 1T & 63.9 & 59.7 & 42.4 & 19.3 & 13.9 \\
    \methodname\ (Ours) & 1T & 63.1 & 57.9 & 42.8 & 18.5 & 12.0 \\
\midrule
    Dense               & 130B & 54.1 & 41.5 & 33.0 & 12.2 & 2.7 \\
    Reg. MoE            & 130B & 60.1 & 51.0 & 37.5 & 15.8 & 5.2 \\
    \methodname\ (Ours) & 130B & 59.1 & 49.2 & 38.1 & 15.5 & 4.2 \\

\bottomrule
\end{tabular}
\vspace{.7em}
\caption{
    \textbf{Full-model Evaluation (\S\ref{subsec:full_model_eval})}. All models are trained on the same data mixture, and activate the same number of parameters (1B). 
    \methodname matches the performance of a standard MoE. \\
    {\small $\dagger$: Use the outdated architecture (no pre-norm, use QK-norm, no shared expert, micro-batch load balancing) and has 64 total experts instead of 128.}
}
\label{tab:full_model_evals}
\end{table}

%% file: figures/modular_use_1t.tex
\begin{figure*}[t]
\centering
\includegraphics[width=1\linewidth]{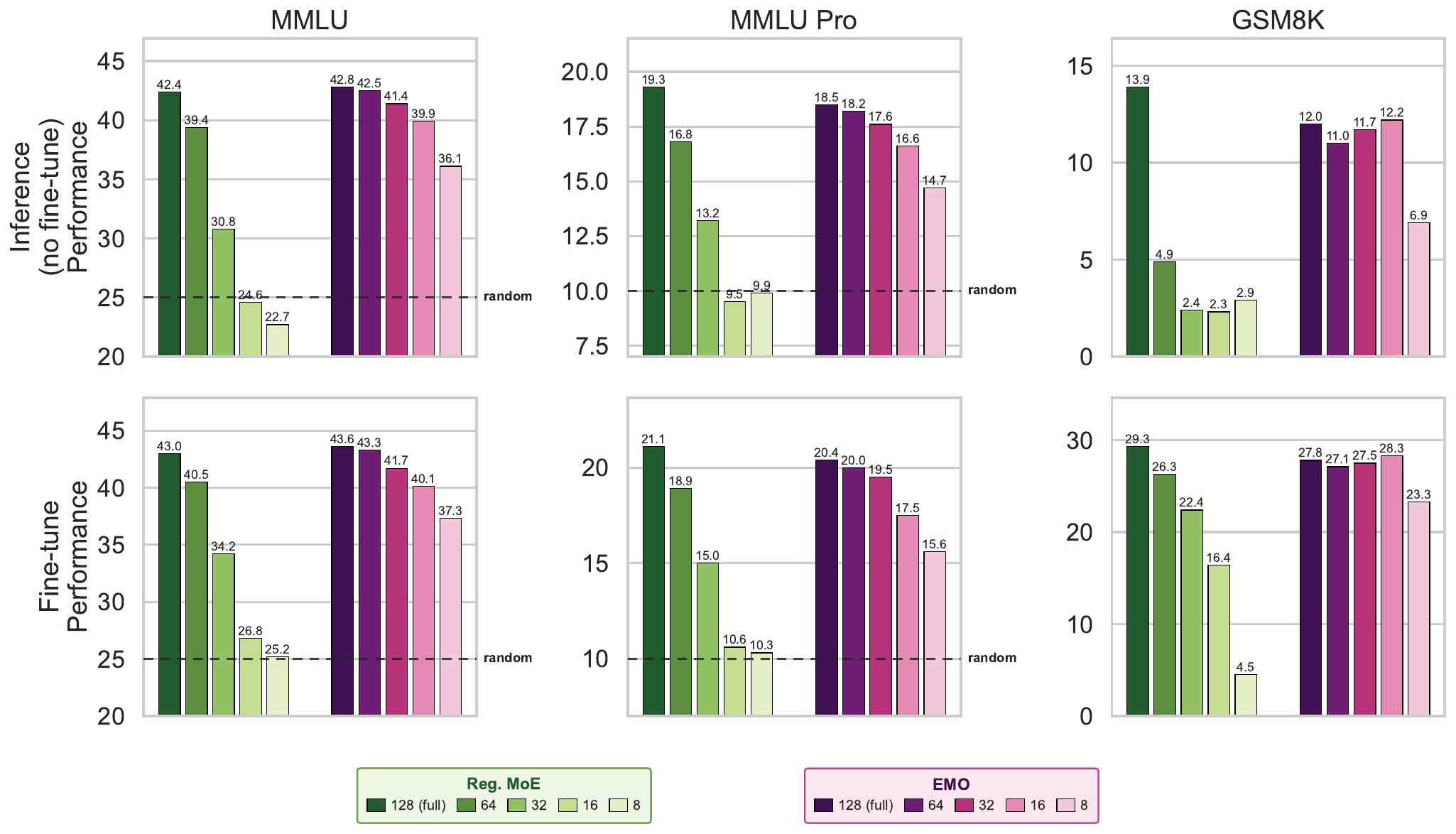}
\vspace{-1.5em}
\caption{\textbf{\SELECTIVEEVAL} of MoEs trained on \textbf{\emph{1T}} tokens (\S\ref{subsec:selective_eval}). Results are shown both without fine-tuning (\emph{top}) and with fine-tuning (\emph{bottom}).
For MMLU and MMLU-Pro, each domain selects a corresponding expert subset as described in \S\ref{subsec:eval}, and we report macro-averaged results across domains (16 for MMLU and 13 for MMLU-Pro).
The baseline MoE degrades sharply under subset restriction, whereas \methodname\ remains robust, with $\approx1$\% drop at 25\% parameters and $\approx3$\% drop at 12.5\%, in both without and with fine-tuning.
}
\vspace{-.5em}
\label{fig:modular_use_1t}
\end{figure*}

%% file: figures/expert_selection_method.tex
\begin{figure*}[t]
\centering
\includegraphics[width=1\linewidth]{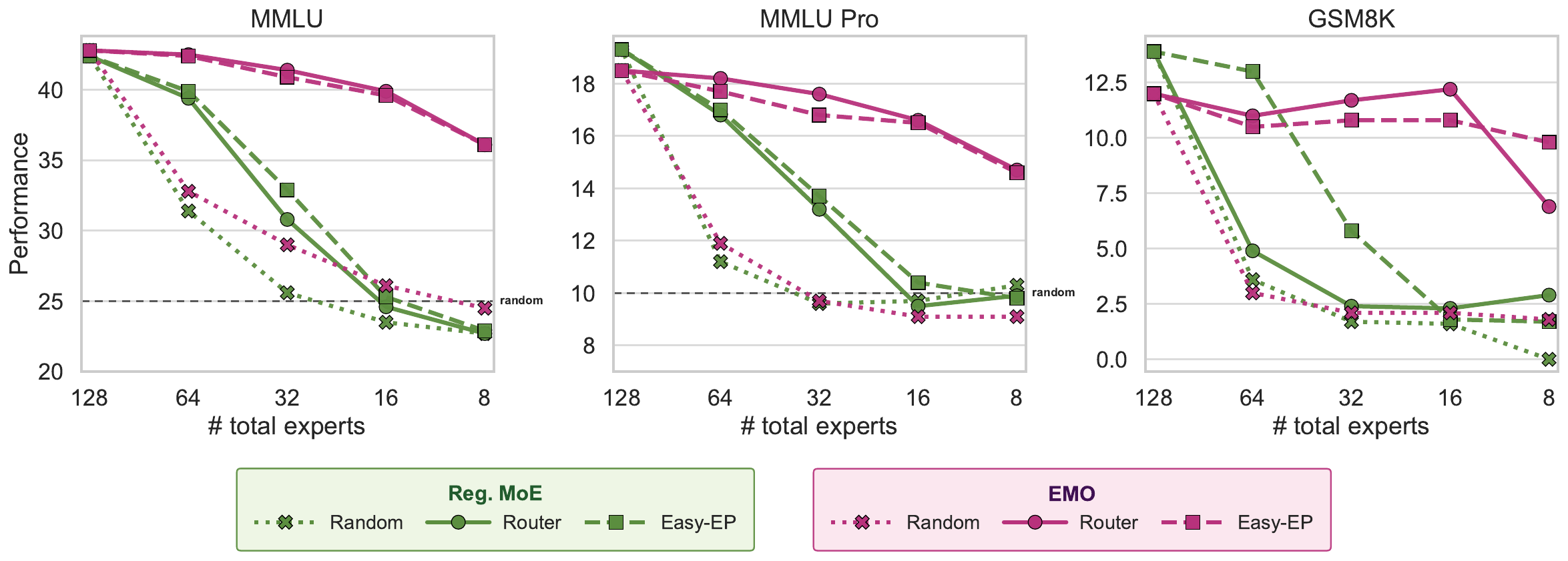}
\vspace{-1.5em}
\caption{
\textbf{Expert Selection Methods} in \selectiveeval\ of MoEs trained on \emph{1T} tokens (\S\ref{subsec:selective_eval}). Results are without fine-tuning.
For MMLU and MMLU-Pro, each domain selects a corresponding expert subset as described in \S\ref{subsec:eval}. 
\methodname expert subsets maintain high performance compared to regular MoEs across both router-based and Easy-EP expert-selection strategies. Random expert selection converges quickly to random performance as expert subsets shrink. 
}
\vspace{-.5em}
\label{fig:expert_selection_method}
\end{figure*}

%% file: figures/token_cluster_comparison.tex
\begin{figure*}[t]
\centering
\includegraphics[width=1\linewidth]{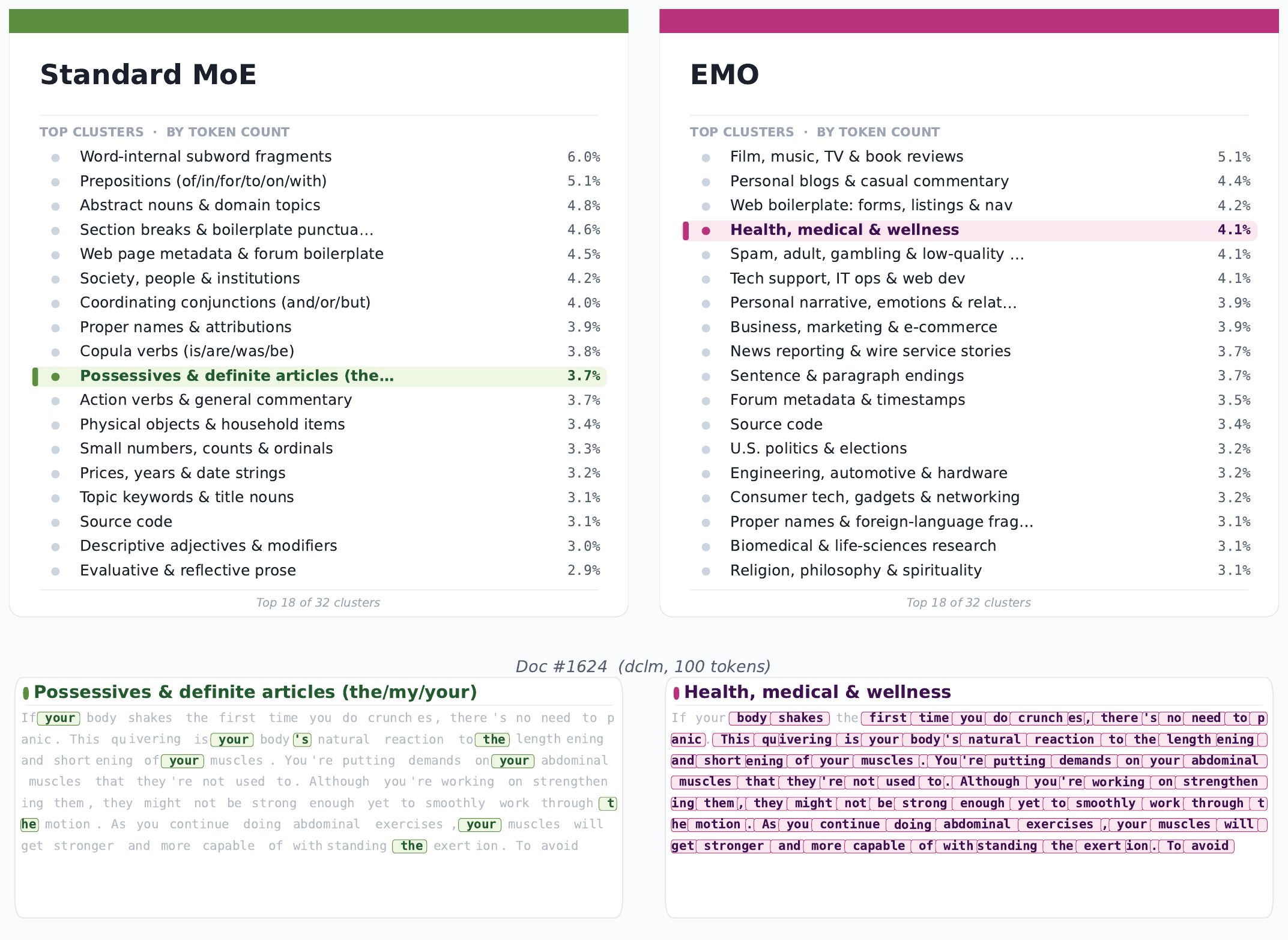}
\vspace{-.5em}
\caption{
\textbf{Token Clusters} of pretraining data on MoEs trained on \emph{1T} tokens, clustered according to the process described in \S\ref{subsec:semantic_specialization}. Claude Code was used to assign a representative short description for each cluster. 
\methodname clusters correspond to semantically meaningful domains, with tokens from the same document largely grouped together. Standard MoE training produces clusters of surface-level or syntactic features, with document tokens dispersed across multiple clusters.}

\vspace{-.5em}
\label{fig:token_cluster_comparison}
\end{figure*}

%% file: figures/domain_similarities.tex
\begin{figure*}[t]
\centering
\includegraphics[width=1\linewidth]{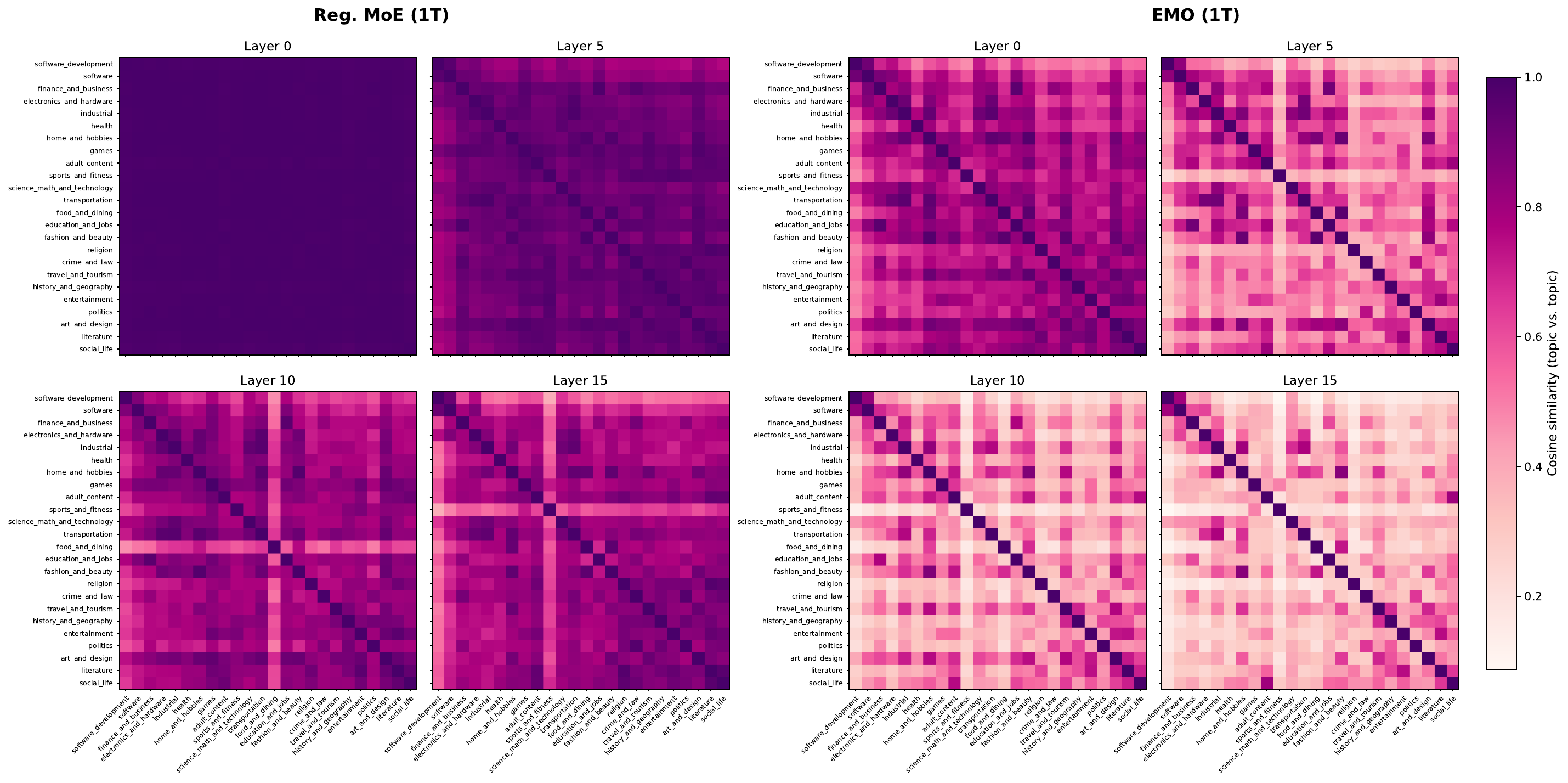}
\vspace{-.5em}
\caption{
\textbf{Domain Similarity} of WebOrganizer documents on MoEs trained on \emph{1T} tokens (\S\ref{subsec:semantic_specialization}). Expert utilization is highly similar (similarity $> 0.6$ for most domain pairs) in regular MoEs, while they are much more distinguishable (similarity $<0.4$) in \methodname, especially in later layers. 
}
\vspace{-.5em}
\label{fig:domain_similarities}
\end{figure*}

%% file: sections/07_discussion.tex
\methodname is among the first MoE models with emergent modularity. While this work primarily focuses on modularity for efficient deployment, it also points to several broader opportunities.

\paragraph{Accessible Deployment of Large Sparse MoEs.}
As MoE models scale to trillions of parameters \cite{deepseekai2026deepseekv4, kimiteam2026kimik2openagentic}, deploying or adapting them becomes increasingly resource-intensive. Prior work addresses this through memory-constrained scaling and optimized serving \cite{li2026can, song2025blockffnendsideaccelerationfriendlymixtureofexperts, shen2026temporally, tairin2025emoe}. Modularity offers an orthogonal path: selectively using small subsets of experts for a given domain, enabling more accessible deployment and adaptation, particularly well-suited for large, highly sparse models.

\paragraph{Fine-grained Control.} Modularity can enable finer-grained control at inference time. Since \methodname organizes experts along semantic domains, subsets could be selectively enabled or disabled depending on the application. 
For example, clusters associated with spam, gambling, or adult content (Figure~\ref{fig:token_cluster_comparison}) can be excluded in child-facing applications. Similarly, specialized domains, e.g., biomedical knowledge, may be valuable for benign use but risky in misuse scenarios, and could be conditionally exposed.

This suggests a potential alternative to dataset-level filtering: isolating and managing capabilities at inference time depending on scenarios.

\paragraph{Modular Development and Maintenance.}
Modularity can also open up a new paradigm for model updates.
Current language models are usually trained and maintained as one large system, requiring the full model, data, and compute to be available at once. A modular model could instead support modular pretraining: training task or domain-specific subset of experts and then reintegrating those experts into the full model. 

As a preliminary test, we finetune a 32-expert subset from \methodname, then inserted it back into the original 128-expert model by replacing the corresponding experts. 
The resulting model improves over the original full model, though it does not yet match the performance of the standalone subset. This provides early evidence that expert subsets can be trained independently and later integrated, which we leave to future work.

\paragraph{Higher Degrees of Monitorability.} Finally, modularity can also make models easier to monitor and audit. Expert activations provide a structured signal of which parts of the model are being used for a given input. For example, if a model answers a math question while strongly activating an expert subset associated with creative writing or low-quality web content, that mismatch may warrant closer inspection. This gives model developers a more structured interface for understanding and debugging model behavior.

%% file: sections/conclusion.tex
We introduced \methodname, a mixture-of-experts model designed to make modularity emerge during pretraining. By constraining tokens within the same document to route through a shared expert pool, \methodname induces expert subsets that specialize to high-level tasks and capabilities without relying on human-defined domains or task labels. Our results show that this structure does not come at the cost of general performance: as a full model, \methodname matches standard MoEs, while its extracted expert subset remain effective even when only a small fraction of experts are retained. Beyond efficient deployment, our analyses show that \methodname learns expert subsets aligned with semantic domains rather than surface-level token patterns, suggesting a qualitatively different form of specialization. Together, these results demonstrate that large language models need not remain monolithic systems. Modularity can be built into pretraining itself, opening a path toward models that are easier to deploy, adapt, inspect, and compose.

%% file: sections/06_appendix.tex
\section{Architectural \& Training Details and Ablations}\label{appendix:hyperparameters}

\input{figures/global_vs_local_lb}

\subsection{Load Balancing}\label{subsec:lb_implementations}

We begin by making explicit how the simplified formulation of load balancing as described in Section \ref{subsec:moe_obj} is implemented in practice.

Under standard implementations, the statistics $\bar{f}_i$ and $\bar{P}_i$ are computed independently within each data parallel group. Let there be $n_p$ data parallel groups\footnote{In this work, we pre-train with data parallelism only.}. For each group $j$, let $f_i^j$ denote the fraction of tokens in its micro-batch routed to expert $i$, and let $P_i^j$ denote the average routing probability assigned to expert $i$ across tokens in that micro-batch. The load balancing loss is then defined as
\[
\mathcal{L}_{\mathrm{LB}} = \frac{1}{n_p} \sum_{j=1}^{n_p} \left[ n_r \sum_{i=1}^{n_r} f_i^j \cdot P_i^j \right],
\]
which corresponds to computing the simplified objective separately within each group and averaging the results.

In this paper, we instead modify the load balancing objective to operate over aggregated routing statistics across data parallel groups. Specifically, we perform an all-reduce over data parallel groups to obtain the aggregated routing frequency
\[
\bar{f}_i = \sum_{j=1}^{n_p} f_i^j,
\]
while retaining the per-group routing probabilities $P_i^j$. The load balancing loss is then computed as
\[
\mathcal{L}_{\mathrm{LB}}
= \frac{1}{n_p} \sum_{j=1}^{n_p} \left[ n_r \sum_{i=1}^{n_r} \bar{f}_i \cdot P_i^j \right].
\]

This formulation, proposed and used in Qwen 3 \cite{yang2025qwen3technicalreport, qiu2025demonsdetailimplementingload}, replaces local (micro-batch-level) estimates of $\bar{f}_i$ with aggregated statistics across data parallel groups, yielding a closer approximation to global routing behavior. In our setting, where training uses data parallelism only, this corresponds to computing load balancing over a larger set of sequences (i.e., the global batch up to gradient accumulation).

We find that global load balancing is critical for stable training in \methodname. As shown in Figure~\ref{fig:global_vs_local_lb}, models trained with standard micro-batch-level load balancing exhibit unstable behavior, while global load balancing leads to consistent and reliable training dynamics. This difference arises because micro-batch-level load balancing conflicts with our document-level routing constraint by enforcing uniform expert usage within each micro-batch. In contrast, global load balancing aggregates routing statistics across data-parallel groups, allowing expert utilization to diversify across documents while preserving consistent routing within each document.

\input{tables/dynamic_d}

\subsection{How to Choose $d$}

We find that varying the size $d$ during training is important for enabling flexible \selectiveeval\ usage at inference time. As shown in Table~\ref{tab:shared_dynamic_ablation}, models trained with a fixed $d$ perform well at that specific expert subset size but degrade when evaluated at other subset sizes. In contrast, training with a distribution over $d$ yields robust performance across a wide range of expert subset sizes.

\subsection{Shared Experts}

In Table~\ref{tab:shared_dynamic_ablation}, we find that incorporating shared experts improves performance in \methodname. This is consistent with prior observations in DeepSeek-MoE \cite{dai2024deepseekmoeultimateexpertspecialization}.

\subsection{Tuning LR and LB} \label{subsec:lr_lb_ablations}

\input{figures/moe_hyperparameter_ablations}

Due to limited compute budget, we perform ablations first on learning rate, then on load balancing in a sequential manner. Furthermore, we ablate the coefficients of each in increments of 10x. 

\paragraph{Standard MoE Hyperparameter Ablations.} In Figure~\ref{fig:moe_hyperparameter_ablations}, we initialize with the default lr = $4e-4$, lb = $1e-2$ configurations following OLMoE \cite{muennighoff2025olmoeopenmixtureofexpertslanguage}. We first ablate the learning rate by increasing it to $4e-3$ and $4e-2$ while keeping the load balancing coefficient fixed. In our results, using a learning rate of $4e-3$ led to the best results. We then ablated the load balancing coefficient from $1e-2$ to $1e-1$, which helped training stablility. We ended up selecting the hyperparameter configurations of lr = $4e-3$ and lb = $1e-1$.

\input{figures/twolevel_hyperparameter_ablations}
\paragraph{\methodname Hyperparameter Ablations. } In Figure~\ref{fig:twolevel_hyperparameter_ablations}, we train \methodname immediately using lb = $1e-1$ and ablated learning rate across lr = $4e-4$ (OLMoE default), $4e-3$, and $4e-2$. We noticed that lr = $4e-3$ led to the strongest performance by 3000 training steps and decided to move forwards with the final configuration of lr = $4e-3$ and lb = $1e-1$. We noticed small spikes in the loss during training for lr = $4e-3$, which were reduced when we used a global version of load balancing, see \ref{subsec:lb_implementations}.

\subsection{Prenorm vs ReorderedNorm}\label{subsec:prenorm_ablation}

\input{figures/prenorm_noqknorm_ablations}

Both \methodname and the standard MoE in this work implemented Prenorm with removed QK-norm instead of the default ReorderedNorm implementation in OLMoE \cite{muennighoff2025olmoeopenmixtureofexpertslanguage}. We ran ablations that consistently showed that our new implementation achieves a lower training loss, see Figure~\ref{fig:prenorm_noqknorm_ablations}.

\section{\Selectiveeval\ Details}

\subsection{130B Token Experiments}\label{appendix:selective_130B_ablation}
\input{figures/modular_use_130b}

Following \S\ref{subsec:selective_eval}, we show the results for \selectiveeval across MMLU, MMLU Pro, and GSM8K for models trained on 130B tokens. Additionally, we compare against memory-matched models traine from scratch, including a standard MoE with 32 experts and a dense model. A subset of these results are presented in Figure~\ref{fig:eval_flowchart} (right). \methodname match or outperform all baselines in \selectiveeval, pushing the pareto-frontier in memory-accuracy trade-off.  

\subsection{Ablations on expert subset initialization}\label{appendix:validation_set_ablations}

\input{figures/calibration_sample_ablation}

We now investigate how validation data affects expert selection. To study this, we vary three factors in Figure~\ref{fig:calibration_sample_ablation}: (1) the number of validation examples used for expert selection, (2) validation set data format (few-shot vs. zero-shot prompts), and (3) the test-set data format (few-shot vs. zero-shot prompts).

\paragraph{\methodname is Sample-efficient in Expert-selection. }

We first consider the default setting used in this work, when both validation and test set use few-shot prompts. In this setting, \methodname\ shows little degradation as validation set size decreases—even down to a single example (red in Figure~\ref{fig:calibration_sample_ablation}. We hypothesize this robustness arises because of the presence of few-shot demonstrations in each validation datapoint, which may provide sufficient token-level signals.

We then investigate using zero-shot prompts for validation and evaluation (gray), performance degrades slightly as the validation set size decreases, but the drop remains modest even with only 5 validation examples. Overall, this demonstrations that \methodname\ remains effective even in highly data-constrained settings.

\paragraph{\methodname\ Depends on Validation Data.}
While \methodname is sample-efficient, the relationship between validation data and expert subset performance is nuanced and task-dependent. On GSM8K, for example, performance improves as the validation set size decreases. One possible explanation is that smaller validation sets produce more focused estimates of expert relevance, whereas aggregating across multiple examples can smooth these signals and yield less specialized expert subsets. 

We also find that validation format sometimes matters independently of evaluation format: selecting experts using few-shot prompts can outperform zero-shot prompts, even when test examples are evaluated using zero-shot formats. This suggests that both the content and structure of validation data play a key role in shaping expert selection, an interaction we leave to future work.

\subsection{\SELECTIVEEVAL\ on "Other" Category in MMLU and MMLU Pro}\label{appendix:mmlu_other_ablation}
\input{figures/modular_use_mmlu_other_130b}

When the deployment task is general (e.g MMLU other and MMLU Pro categories, which serve as a "catch-all" for MMLU subjects), \methodname expert subsets of size 32 and 8 experts struggle to match the Reg MoE @ 32 and Dense @8 baseline models trained from scratch (Figure~\ref{fig:modular_use_mmlu_other_130b}). We view this phenomenon as a property of modular models, and believe it provides concrete evidence that \methodname works in \selectiveeval\ because it has groups of experts that have localize capabilities. When reporting aggregate metrics of MMLU and MMLU Pro, we intentionally exclude including the "other" category, as it is not an example of \selectiveeval\ on a specific task. 

\subsection{Annealing Standard MoEs to be Modular.}\label{appendix:modular_anneal} 
\input{tables/anneal_ablation}
We investigate whether modularity requires applying the document-level expert pool objective (\S\ref{subsec:ours_obj}) throughout pre-training, or whether a standard MoE can be made modular after pre-training. To test this, we take a standard MoE pretrained on 1T tokens and anneal it using the document-level expert pool objective instead of the standard MoE training objective. Denoted as \methodname-anneal in Table~\ref{tab:modmoe_anneal_vs_modmoe}, this model underperforms \methodname\ on most benchmarks across most expert subset sizes. However, \methodname-anneal still trains successfully and exhibits signs of modularity, suggesting that post-training applications of the document-level expert pool objective may be a promising direction for future work.

\subsection{Generations from \SELECTIVEEVAL\ on GSM8K} \label{appendix:gsm8k_examples}

We now demonstrate how small expert subsets of \methodname is qualitatively better than that of regular MoEs. We give examples of GSM8K generations of \methodname and Regular MoE trained on 1T tokens under \selectiveeval\ across different expert subset sizes. No finetuning was performed (expert subsets are evaluated zero-shot). We note that 8-expert subsets of \methodname consistently produces coherent outputs while regular MoEs subsets deteriorate.

\input{tables/gsm8k_examples}

\section{Evaluation Details}\label{appendix:evaluation_detail}

\input{tables/eval_task_details}

\paragraph{MMLU and MMLU-Pro Setup.}
For MMLU and MMLU-Pro, we randomly sample 40\% of examples for the validation/training set and use the remaining 60\% for evaluation. For MMLU, we group the original 57 subjects into 17 broader categories following~\citep{hendryckstest2021} to ensure sufficient examples per category to train and evaluate with. Expert selection is performed at the category level: all subjects within the same category are processed by a single group of experts. When reporting aggregate MMLU and MMLU-Pro results, we exclude the ``other'' category, which is discussed in Appendix~\ref{appendix:mmlu_other_ablation}.

\paragraph{Evaluation and Finetuning Protocol.}
We provide the full list of evaluation categories and subjects in Table~\ref{tab:eval_tasks}. For multiple-choice benchmarks, we score each answer choice by its log-likelihood and select the highest-scoring option, reporting raw accuracy (``acc-raw''). For generation benchmarks, we report recall on Gen5 and exact match on \textsc{GSM8K}.

During \selectiveeval, we use the same set of examples as both the validation data for expert selection and training data for finetuning. For tasks other than MMLU and MMLU-Pro, we merge the original train and validation splits and use the combined set for both expert selection and finetuning. When finetuning is performed, we mask the input portion of the prompt and optimize only over output tokens. Unless otherwise specified, finetuning follows standard Hugging Face settings with one epoch, batch size 32, and learning rate $5\times 10^{-5}$.

\section{Token Clustering Details}
\paragraph{\methodname forms a Distinct Cluster for the First Token of Each Document.}
We observe a dedicated expert cluster consistently activated for the first token in each document. This behavior is intuitive, as the model processes the first token without any context. Interestingly, this cluster does not extend to the second token: in most cases after observing just one token, the model transitions into a specific expert activation pattern that often remains stable throughout the rest of the document. This suggests that \methodname rapidly commits to a document-level routing pattern after minimal context, with the first token serving as a distinct initialization phase before more specialized processing begins.

%% file: figures/global_vs_local_lb.tex
\begin{figure}[H]
\centering
\includegraphics[width=.6\linewidth]{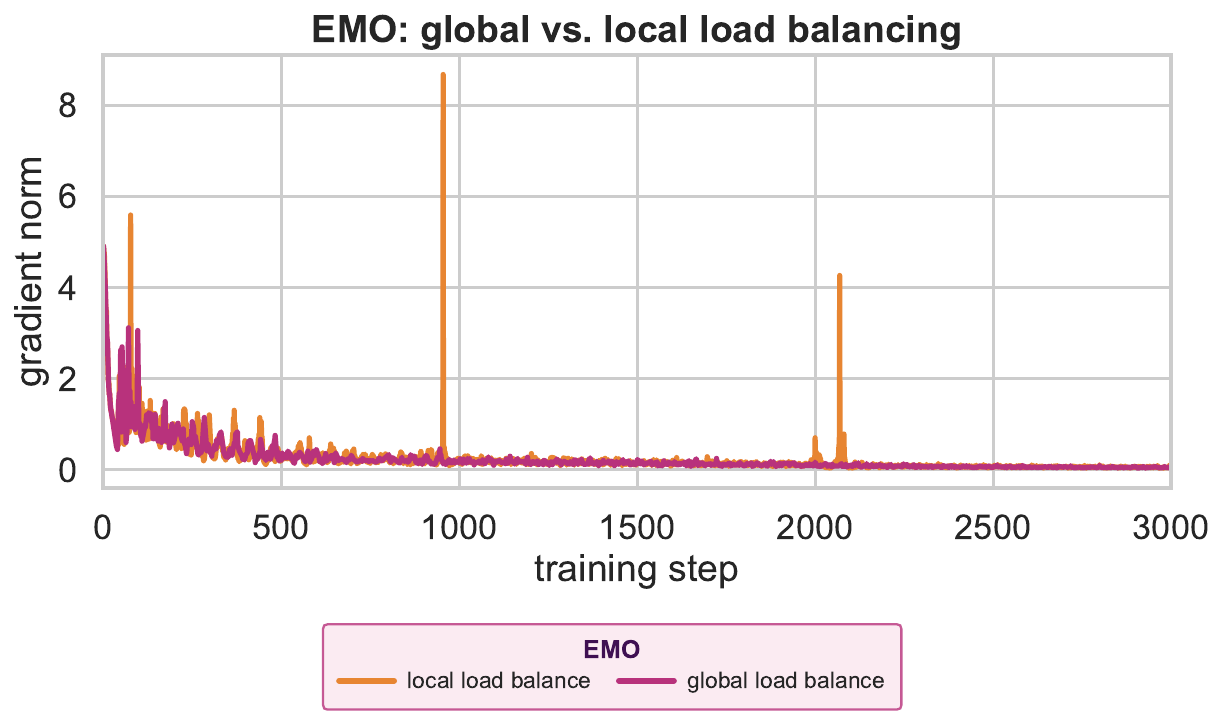}
\vspace{-.5em}
\caption{\textbf{Global vs Local Load Balancing} and its effects on training stability (Appendix~\ref{subsec:lb_implementations}). Using global load balancing leads to more stable pre-training runs with less gradient norm spikes. }
\vspace{-.5em}
\label{fig:global_vs_local_lb}
\end{figure}

%% file: tables/dynamic_d.tex
\begin{table}[H]
\centering
\footnotesize
\setlength{\tabcolsep}{4.5pt}
\begin{tabular}{ccc ccccc ccccc}
\toprule
\multicolumn{3}{c}{\textbf{Configuration}}
& \multicolumn{5}{c}{\textbf{Inference}}
& \multicolumn{5}{c}{\textbf{Fine-tuning}} \\
\cmidrule(lr){1-3} \cmidrule(lr){4-8} \cmidrule(lr){9-13}
\textbf{$n_r$} & \textbf{$n_s$} & \textbf{$d$}
& \textbf{8} & \textbf{16} & \textbf{32} & \textbf{64} & \textbf{128}
& \textbf{8} & \textbf{16} & \textbf{32} & \textbf{64} & \textbf{128} \\
\midrule
128 & 0 & 32
& 30.2 & 35.3 & 37.2 & 36.0 & 31.9
& 31.7 & 36.2 & 37.8 & 38.4 & 37.3 \\
127 & 1 & 32
& 29.6 & 34.4 & 36.6 & 35.6 & 33.6
& 31.7 & 36.0 & 38.3 & 38.5 & 37.4 \\
127 & 1 & $U(8,128)$
& 33.7 & 36.4 & 37.0 & 37.7 & 38.1
& 34.5 & 37.5 & 38.5 & 39.7 & 39.8 \\
\bottomrule
\end{tabular}
\vspace{1em}
\caption{\textbf{Ablating Shared Experts and Document Pool Size}. Evaluation on MMLU across different architectural configurations, comparing the effects of shared experts and dynamic expert subset sizes. We compare models without a shared expert ($n_s=0$), with a shared expert and fixed expert subset size ($d=32$), and with a shared expert and dynamic expert subset size ($U(8,128)$). Using a shared expert improves performance, while dynamic expert-subset-size training enables more flexible expert selection across different budgets. }
\label{tab:shared_dynamic_ablation}
\end{table}

%% file: figures/moe_hyperparameter_ablations.tex
\begin{figure}[H]
\centering
\includegraphics[width=.6\linewidth]{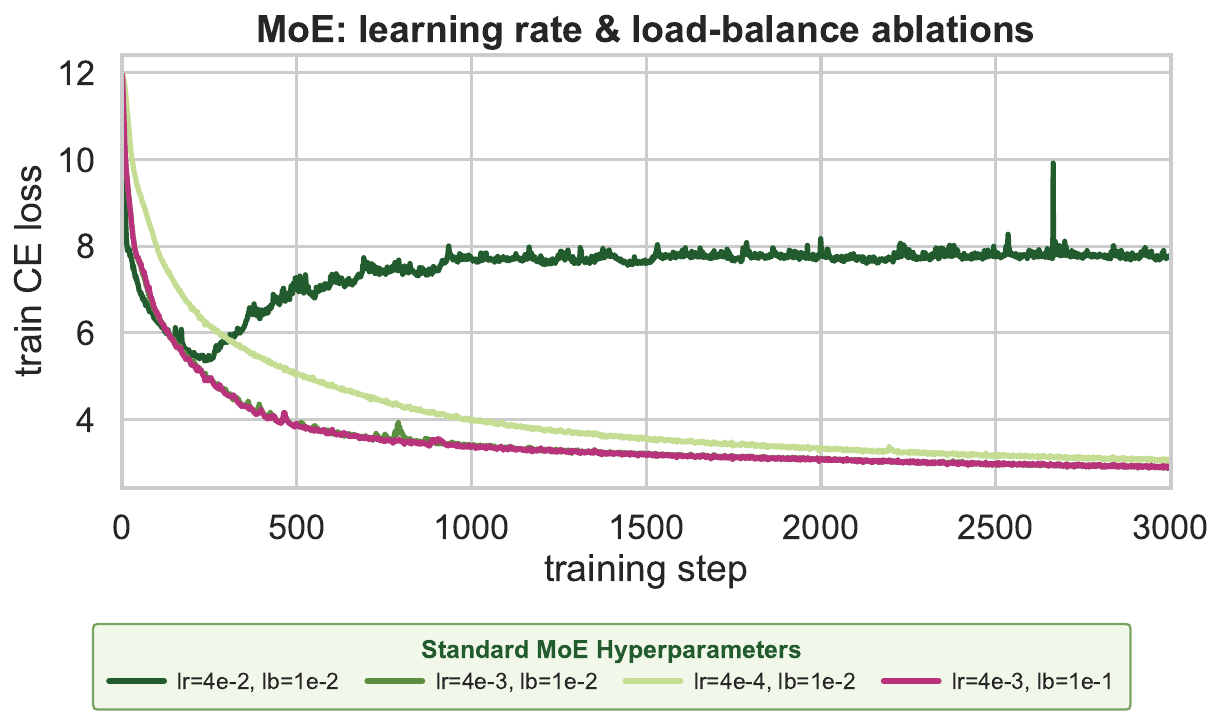}
\vspace{-.5em}
\caption{\textbf{Standard MoE LR and LB Ablations}. We ablate standard MoEs across learning rates and load balancing (Appendix~\ref{subsec:lr_lb_ablations}). We identify the best configuration as lr = $4e-3$ and lb = $1e-1$. For load balancing coefficient, we do not observe significant differences between $1e-1$ and $1e-2$, and choose the former because it had slightly higher training stability. }
\vspace{-.5em}
\label{fig:moe_hyperparameter_ablations}
\end{figure}

%% file: figures/twolevel_hyperparameter_ablations.tex
\begin{figure}[H]
\centering
\includegraphics[width=.6\linewidth]{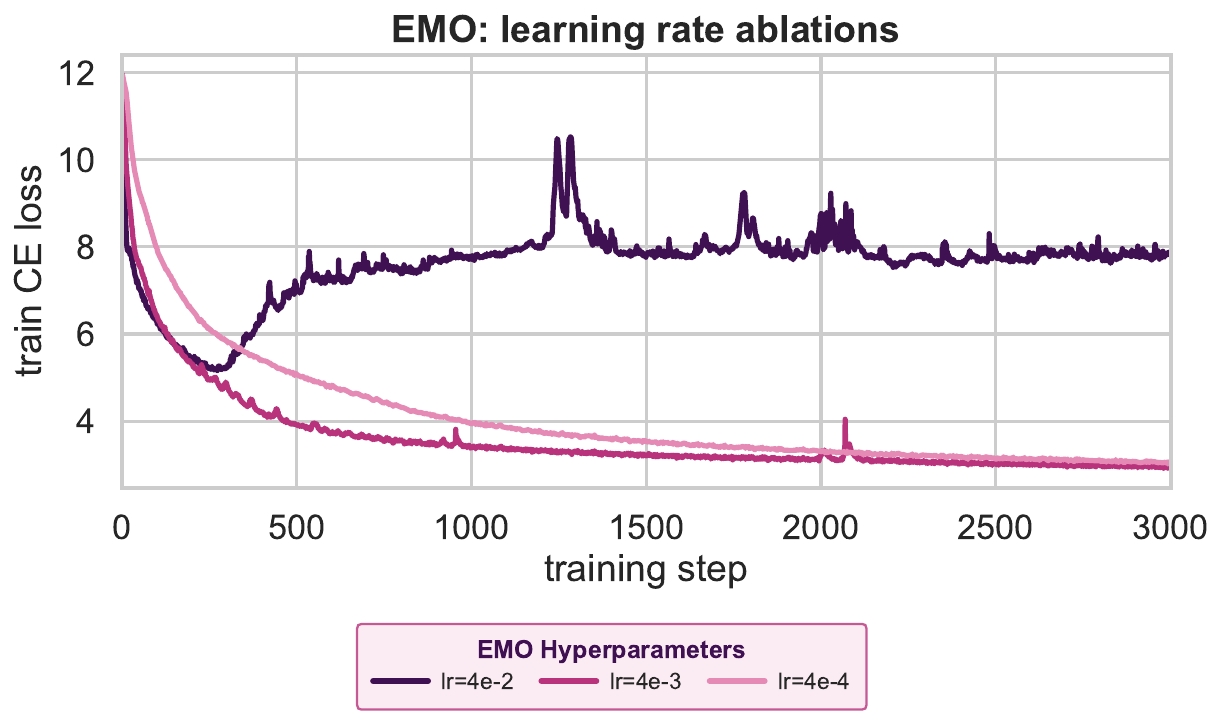}
\vspace{-.5em}
\caption{\textbf{\methodname ablations over LR}. Ablations of \methodname across learning rates. Due to limited compute resources, we fix lb = $1e-1$ and only ablate the learning rate, finding that lr = $4e-3$ offered the best training loss. Minor training loss spikes in lr = $4e-3$ were resolved by implementing load balancing over global batches. }
\vspace{-.5em}
\label{fig:twolevel_hyperparameter_ablations}
\end{figure}

%% file: figures/prenorm_noqknorm_ablations.tex
\begin{figure}[H]
\centering
\includegraphics[width=.6\linewidth]{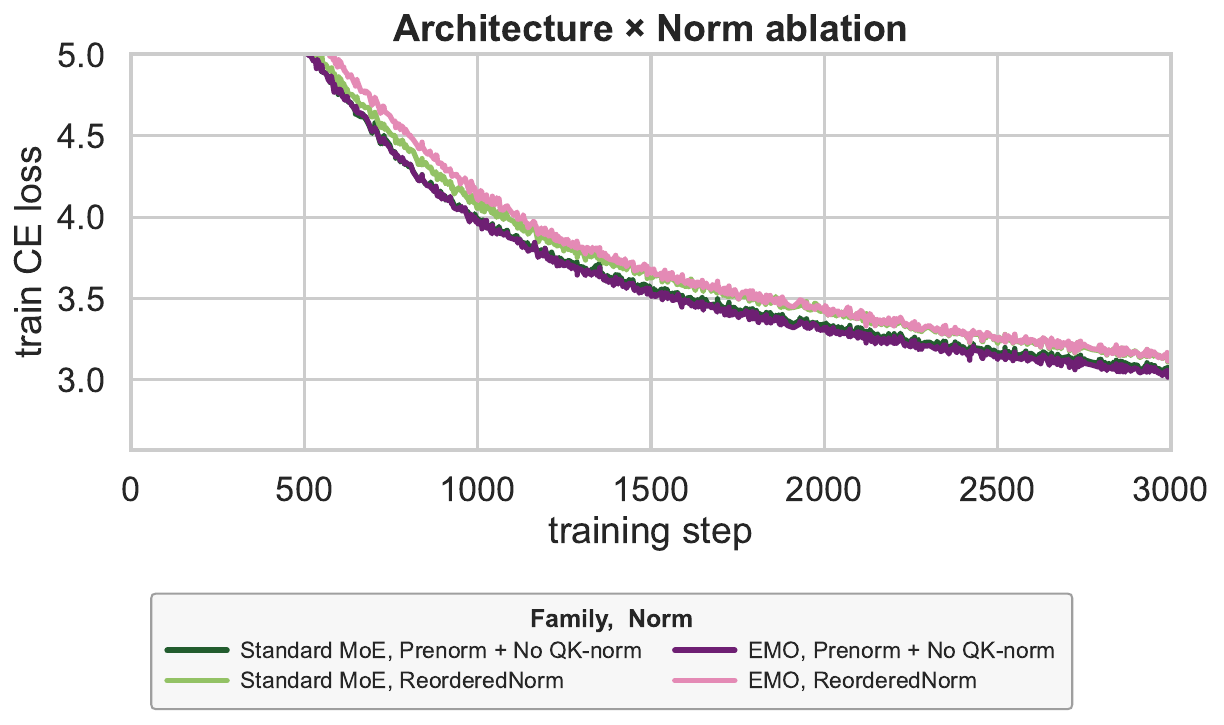}
\vspace{-.5em}
\caption{
\textbf{Prenorm w. No QK Norm Ablations}. Ablations on ReorderedNorm from \cite{muennighoff2025olmoeopenmixtureofexpertslanguage} versus Prenorm with removed QK-norm (Appendix~\ref{subsec:prenorm_ablation}). On both standard MoEs and \methodname, using Prenorm with removed QK-norm achieves lower loss than ReorderedNorm. These experiments were conducted without applying global load balancing, shared experts, and dynamic $d$. }
\vspace{-.5em}
\label{fig:prenorm_noqknorm_ablations}
\end{figure}

%% file: figures/modular_use_130b.tex
\begin{figure*}[t]
\centering
\includegraphics[width=1\linewidth]{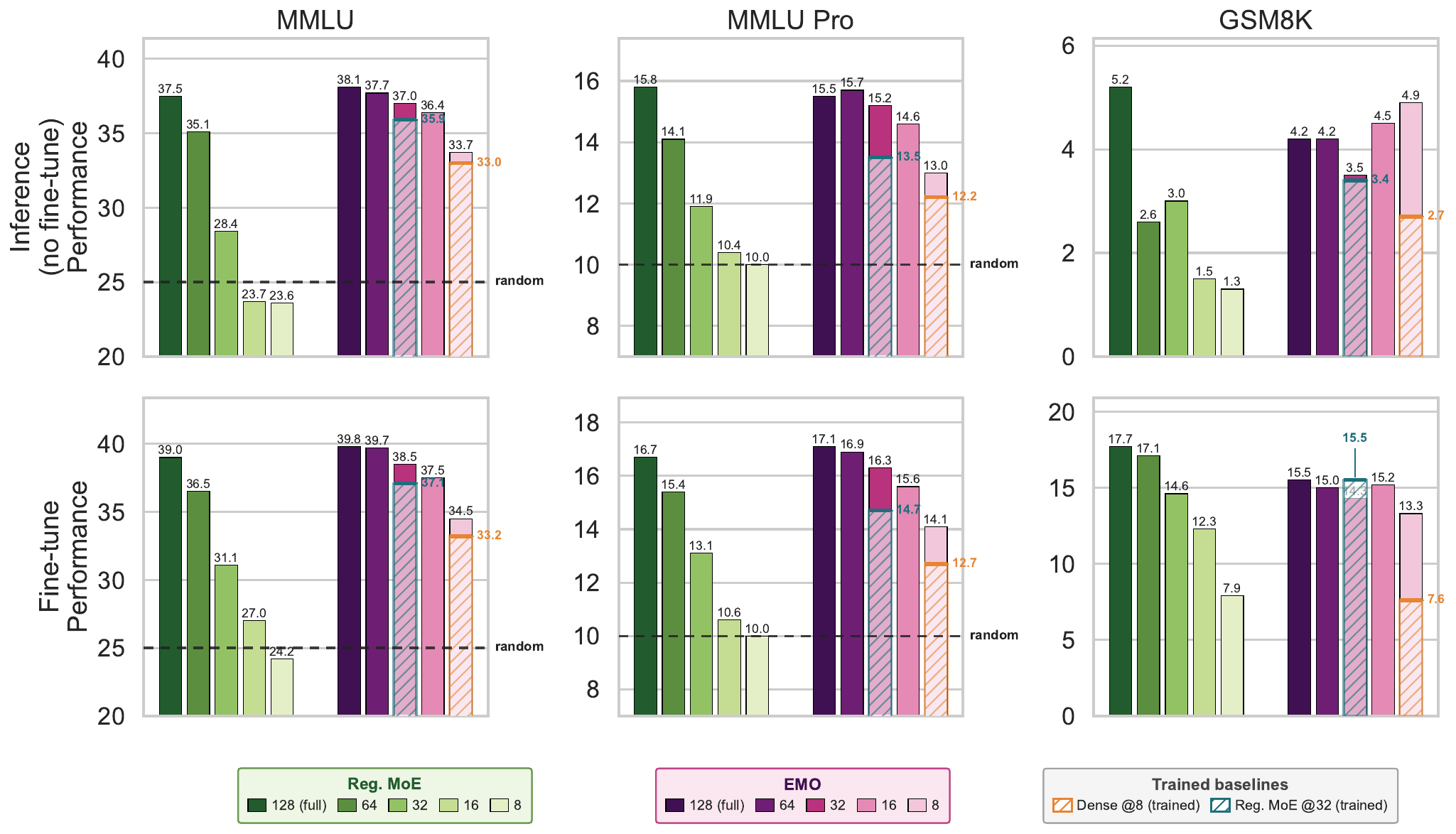}
\vspace{-1.5em}
\caption{
\textbf{\SELECTIVEEVAL} of MoEs trained on \textbf{\emph{130B}} tokens (\S\ref{subsec:selective_eval}). 
We report performance before fine-tuning (\emph{top}) and after fine-tuning (\emph{bottom}). 
For MMLU and MMLU-Pro, each domain selects a corresponding expert subset as described in \S\ref{subsec:eval}, and we report macro-averaged results across domains (17 for MMLU and 14 for MMLU-Pro).
``Trained baseline @$k$'' denotes a model trained from scratch with a parameter count matched to a $k$-expert subset. 
Across all tasks, the \methodname\ 32-expert subset and 8-expert subset match or outperform the corresponding Reg. MoE @ 32 and Dense @ 8 trained models, with the 8-expert subset of \methodname nearly doubling the performance of the Dense @ 8 on GSM8k, both before and after fine-tuning.
}
\vspace{-.5em}
\label{fig:modular_use_130b}
\end{figure*}

%% file: figures/calibration_sample_ablation.tex
\begin{figure}[H]
\centering
\includegraphics[width=1\linewidth]{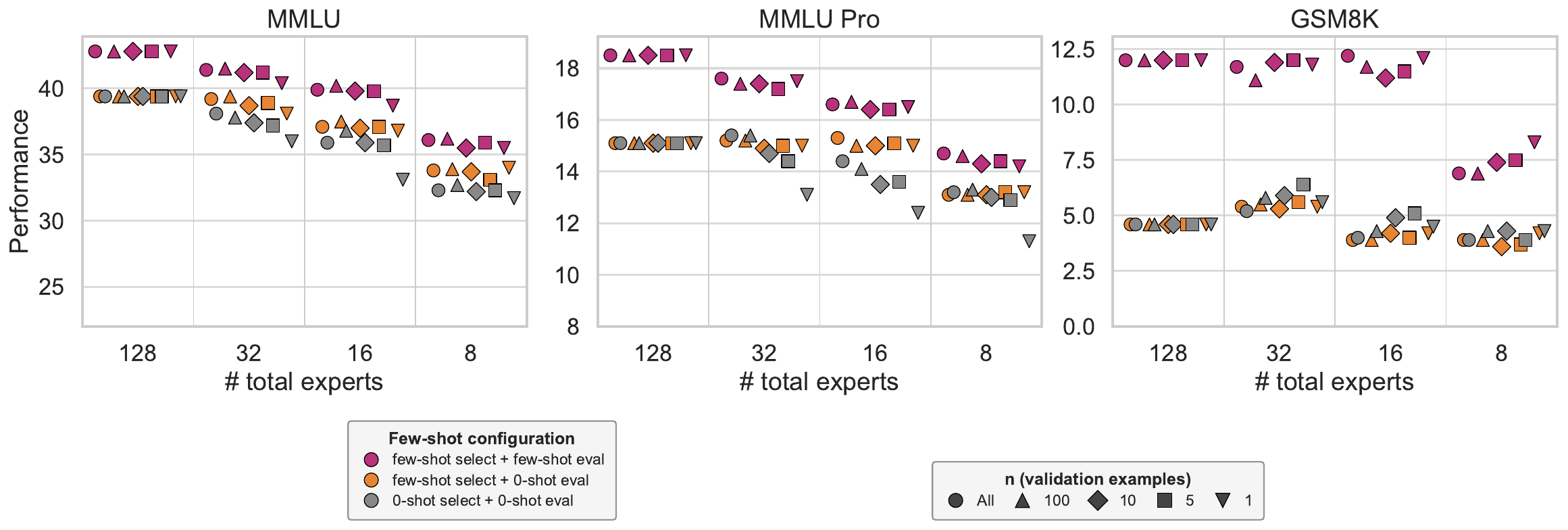}
\vspace{-.5em}
\caption{Effects of \textbf{validation data quantity} ($n$) and the presence of \textbf{few-shot demonstrations} (on both the data used to select experts and the actual test-set queries) on \methodname expert subset performance, without any subsequent fine-tuning. By default MMLU/MMLU Pro have 5-shot demonstrations and GSM8K has 8-shot demonstrations. For GSM8K, we run with three random seeds for n=1, 5, and 10. \methodname is sample-efficient in expert selection, and using few-shot demonstrations during both expert selection and evaluation brings large performance gains. }
\vspace{-.5em}
\label{fig:calibration_sample_ablation}
\end{figure}

%% file: figures/modular_use_mmlu_other_130b.tex
\begin{figure}[H]
\centering
\includegraphics[width=1\linewidth]{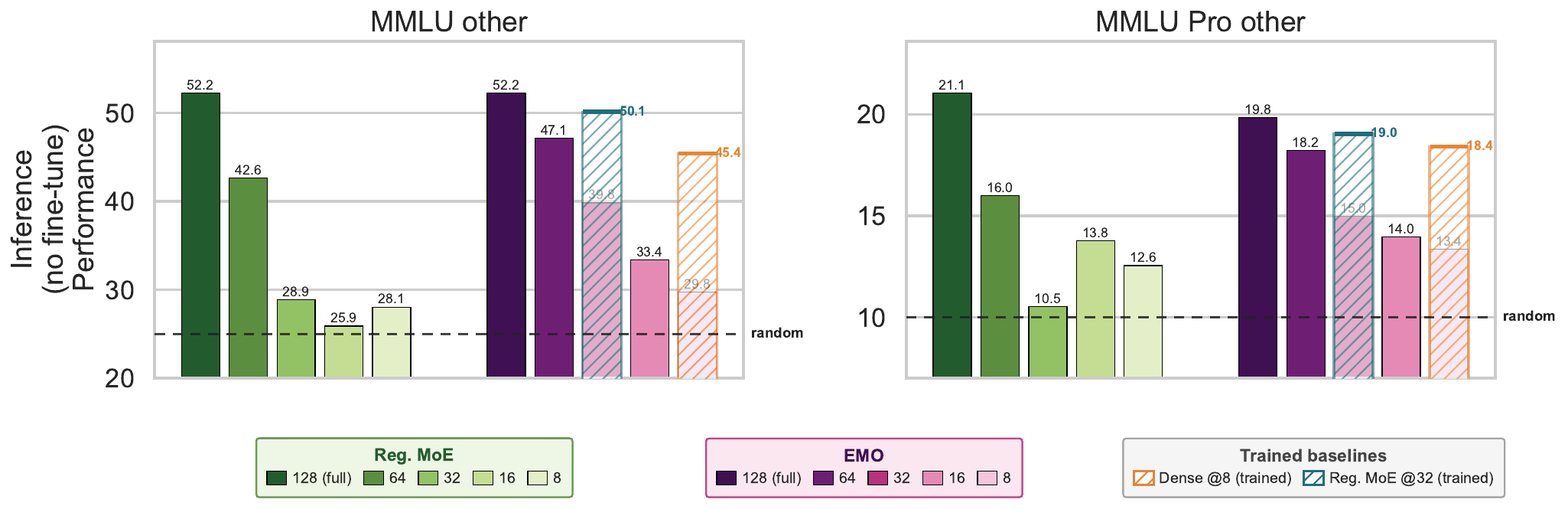}
\vspace{-.5em}
\caption{
\textbf{\Selectiveeval\ on Other category} on \methodname and standard MoEs trained on \emph{130B} tokens (Appendix~\ref{appendix:mmlu_other_ablation}). Results are without fine-tuning. ``Trained baseline @$k$'' denotes a model trained from scratch with a parameter count matched to a $k$-expert subset. 
On tasks that are general, \methodname expert subsets of sizes 32 and 8 performs worse compared to ``Reg MoE @$32$'' and ``Dense @$8$'' that are trained from scratch. }
\vspace{-.5em}
\label{fig:modular_use_mmlu_other_130b}
\end{figure}

%% file: tables/anneal_ablation.tex
\begin{table}[H]
\centering
\footnotesize
\setlength{\tabcolsep}{3.5pt}
\resizebox{\linewidth}{!}{
\begin{tabular}{l l rrr rrr}
\toprule
 & \multirow{2}{*}{\# Experts}
 & \multicolumn{3}{c}{\textbf{Inference}}
 & \multicolumn{3}{c}{\textbf{Fine-tuning}} \\
\cmidrule(lr){3-5} \cmidrule(lr){6-8}
 &
 & \textbf{MMLU}
 & \textbf{MMLU-Pro}
 & \textbf{GSM8K}
 & \textbf{MMLU}
 & \textbf{MMLU-Pro}
 & \textbf{GSM8K} \\
\midrule
\multirow{5}{*}{\methodname-anneal}
    & 8 & 32.1 & 13.1 & 7.3 & 33.5 & 14.2 & 22.6 \\
    & 16 & 35.4 & 14.8 & 9.9 & 37.3 & 16.3 & 25.3 \\
    & 32 & 38.8 & 16.5 & 11.3 & 39.9 & 18.3 & 26.8 \\
    & 64 & 41.3 & 17.4 & 12.8 & 42.7 & 19.5 & 27.7 \\
    & 128 (trained) & 42.3 & 18.2 & 13.0 & 43.7 & 20.1 & 27.2 \\
\midrule
\multirow{5}{*}{\methodname}
    & 8 & 36.1 & 14.7 & 6.9 & 37.3 & 15.6 & 23.3 \\
    & 16 & 39.9 & 16.6 & 12.2 & 40.1 & 17.5 & 28.3 \\
    & 32 & 41.4 & 17.6 & 11.7 & 41.7 & 19.5 & 27.5 \\
    & 64 & 42.5 & 18.2 & 11.0 & 43.3 & 20.0 & 27.1 \\
    & 128 (trained) & 42.8 & 18.5 & 12.0 & 43.6 & 20.4 & 27.8 \\
\bottomrule
\end{tabular}
}
\vspace{.3em}
\caption{
\textbf{Inducing Modularity during Annealing Only.} We compare \methodname, trained from scratch on 1T tokens and annealed on 50B tokens with document-level expert pool constraint (\S\ref{subsec:ours_obj}), against \methodname-anneal, a model trained on 1T tokens as a standard MoE, but annealed on 50B tokens with the document-level expert pool constraint. Across most tasks and expert subset sizes, \methodname improves over \methodname-anneal, indicating that pre-training from scratch is important in realizing gains during \selectiveeval. 
}
\label{tab:modmoe_anneal_vs_modmoe}
\end{table}

%% file: tables/gsm8k_examples.tex
\begin{tcolorbox}[colback=blue!3,colframe=blue!55,boxrule=0.5pt,arc=2pt,left=8pt,right=8pt,top=4pt,bottom=4pt]
\textbf{Q.}\ Harry slept 9 hours last night. His friend James slept only 2/3 of what Harry slept. How many more hours did Harry sleep than James?\hfill\\ \textbf{Gold:}\ \textcolor{goldcol}{\textbf{3}}
\end{tcolorbox}

\vspace{3pt}
\begin{tcbraster}[
  raster columns=2,
  raster equal height=rows,
  raster column skip=5pt,
  raster row skip=4pt,
  fonttitle=\small\bfseries,
  fontupper=\small,
  boxrule=0.4pt,
  arc=2pt,
  left=6pt,right=6pt,top=3pt,bottom=3pt,
]
\begin{tcolorbox}[
  colback=regsoft,colframe=regdark,coltitle=regdark,colbacktitle=regsoft,
  title={Reg.\ MoE, 8-expert subset\hfill \textcolor{nogray}{\ding{55}}}
]
Olivia \$2005 2005 2005 2005 2005 2005 2005 [+162 more] 200
\end{tcolorbox}
\begin{tcolorbox}[
  colback=emosoft,colframe=emomag,coltitle=emodark,colbacktitle=emosoft,
  title={\methodname{}, 8-expert subset\hfill \textcolor{goodgreen}{\ding{51}}}
]
Harry slept 9 hours. His friend James slept 2/3 of 9. So 2/3 of 9 is 6. So the answer is 6 more hours. 9 - 6 = 3. So the answer is 3.
\end{tcolorbox}
\begin{tcolorbox}[
  colback=regsoft,colframe=regdark,coltitle=regdark,colbacktitle=regsoft,
  title={Reg.\ MoE, 32-expert subset\hfill \textcolor{nogray}{\ding{55}}}
]
Harry slept 9 hours last night. So the answer is 9 - 2 = 7. So the answer is 7.
\end{tcolorbox}
\begin{tcolorbox}[
  colback=emosoft,colframe=emomag,coltitle=emodark,colbacktitle=emosoft,
  title={\methodname{}, 32-expert subset\hfill \textcolor{goodgreen}{\ding{51}}}
]
Harry slept 9 hours. James slept 2/3 of 9 hours. So James slept 9 * 2/3 = 6 hours. So Harry slept 9 - 6 = 3 hours more than James.
\end{tcolorbox}
\begin{tcolorbox}[
  colback=regsoft,colframe=regdark,coltitle=regdark,colbacktitle=regsoft,
  title={Reg.\ MoE, 128-expert subset\hfill \textcolor{goodgreen}{\ding{51}}}
]
Harry slept 9 hours. James slept 2/3 of that. So James slept 9 * 2/3 = 6 hours. So Harry slept 9 - 6 = 3 hours more than James.
\end{tcolorbox}
\begin{tcolorbox}[
  colback=emosoft,colframe=emomag,coltitle=emodark,colbacktitle=emosoft,
  title={\methodname{}, 128-expert subset\hfill \textcolor{goodgreen}{\ding{51}}}
]
Harry slept 9 hours. James slept 2/3 of 9 hours. So James slept 9 * 2/3 = 6 hours. So Harry slept 9 - 6 = 3 hours more than James.
\end{tcolorbox}
\end{tcbraster}

\vspace{6pt}

\begin{tcolorbox}[colback=blue!3,colframe=blue!55,boxrule=0.5pt,arc=2pt,left=8pt,right=8pt,top=4pt,bottom=4pt]
\textbf{Q.}\ Tom gets 4 car washes a month.  If each car wash costs \$15 how much does he pay in a year?\hfill \\ \textbf{Gold:}\ \textcolor{goldcol}{\textbf{720}}
\end{tcolorbox}

\vspace{3pt}
\begin{tcbraster}[
  raster columns=2,
  raster equal height=rows,
  raster column skip=5pt,
  raster row skip=4pt,
  fonttitle=\small\bfseries,
  fontupper=\small,
  boxrule=0.4pt,
  arc=2pt,
  left=6pt,right=6pt,top=3pt,bottom=3pt,
]
\begin{tcolorbox}[
  colback=regsoft,colframe=regdark,coltitle=regdark,colbacktitle=regsoft,
  title={Reg.\ MoE, 8-expert subset\hfill \textcolor{nogray}{\ding{55}}}
]
3 4 ? 1000 3 = 9, 5 5 4 + 4 = 9 so 1000 9 = 3 4 = 4 = 5 4 = 5 4 = 5 4 = 5 4 = 5 4 = 5 4 = 5 4 = 5 4 = 5 4 = 5 4 = 5 4 = 5 4 = 5 4 = 5 4 = 5 4 = 5 4 = 5 4 = 5 4 = 5 4 = 5 4 = 5 4 = ...
\end{tcolorbox}
\begin{tcolorbox}[
  colback=emosoft,colframe=emomag,coltitle=emodark,colbacktitle=emosoft,
  title={\methodname{}, 8-expert subset\hfill \textcolor{nogray}{\ding{55}}}
]
Tom started with 4 car washes. Each car wash costs \$15. So 4 * 15 = 60 dollars. So the answer is 60 dollars.
\end{tcolorbox}
\begin{tcolorbox}[
  colback=regsoft,colframe=regdark,coltitle=regdark,colbacktitle=regsoft,
  title={Reg.\ MoE, 32-expert subset\hfill \textcolor{nogray}{\ding{55}}}
]
There were 4 cars. 4 * 15 = 60. So the answer is 60.
\end{tcolorbox}
\begin{tcolorbox}[
  colback=emosoft,colframe=emomag,coltitle=emodark,colbacktitle=emosoft,
  title={\methodname{}, 32-expert subset\hfill \textcolor{goodgreen}{\ding{51}}}
]
Tom gets 4 car washes a month. Each car wash costs \$15. So 4 x 15 = 60. So Tom pays 60 dollars a month. 60 x 12 = 720. So Tom pays 720 dollars a year.
\end{tcolorbox}
\begin{tcolorbox}[
  colback=regsoft,colframe=regdark,coltitle=regdark,colbacktitle=regsoft,
  title={Reg.\ MoE, 128-expert subset\hfill \textcolor{goodgreen}{\ding{51}}}
]
Tom gets 4 car washes a month. 4 car washes a month will be 4 x 12 = 48. 48 car washes a year. 48 car washes a year will be 48 x 15 = 720. 720 dollars a year. So the answer is 720.
\end{tcolorbox}
\begin{tcolorbox}[
  colback=emosoft,colframe=emomag,coltitle=emodark,colbacktitle=emosoft,
  title={\methodname{}, 128-expert subset\hfill \textcolor{goodgreen}{\ding{51}}}
]
Tom gets 4 car washes a month. Each car wash costs \$15. So 4 x 15 = 60 dollars. So Tom pays 60 dollars a month. 60 dollars a month for 12 months is 60 x 12 = 720 dollars a year.
\end{tcolorbox}
\end{tcbraster}

\vspace{6pt}

\begin{tcolorbox}[colback=blue!3,colframe=blue!55,boxrule=0.5pt,arc=2pt,left=8pt,right=8pt,top=4pt,bottom=4pt]
\textbf{Q.}\ Raymond and Samantha are cousins. Raymond was born 6 years before Samantha. Raymond had a son at the age of 23. If Samantha is now 31, how many years ago was Raymond's son born?\hfill \\ \textbf{Gold:}\ \textcolor{goldcol}{\textbf{14}}
\end{tcolorbox}

\vspace{3pt}
\begin{tcbraster}[
  raster columns=2,
  raster equal height=rows,
  raster column skip=5pt,
  raster row skip=4pt,
  fonttitle=\small\bfseries,
  fontupper=\small,
  boxrule=0.4pt,
  arc=2pt,
  left=6pt,right=6pt,top=3pt,bottom=3pt,
]
\begin{tcolorbox}[
  colback=regsoft,colframe=regdark,coltitle=regdark,colbacktitle=regsoft,
  title={Reg.\ MoE, 8-expert subset\hfill \textcolor{nogray}{\ding{55}}}
]
Olivia 3 5 = 4 9 4 Xe ? 5 4 3 Xe ? 5 5 4 = 9 so 20v 5 5 5 5 5 5 [+213 more]
\end{tcolorbox}
\begin{tcolorbox}[
  colback=emosoft,colframe=emomag,coltitle=emodark,colbacktitle=emosoft,
  title={\methodname{}, 8-expert subset\hfill \textcolor{nogray}{\ding{55}}}
]
Raymond is 6 years older than Samantha. So Samantha is 31 - 6 = 25. Raymond is 6 years older than Samantha. So Raymond is 31 - 6 = 25 years old. So the answer is 25.
\end{tcolorbox}
\begin{tcolorbox}[
  colback=regsoft,colframe=regdark,coltitle=regdark,colbacktitle=regsoft,
  title={Reg.\ MoE, 32-expert subset\hfill \textcolor{nogray}{\ding{55}}}
]
Raymond was born 6 years before Samantha. So the answer is 6.
\end{tcolorbox}
\begin{tcolorbox}[
  colback=emosoft,colframe=emomag,coltitle=emodark,colbacktitle=emosoft,
  title={\methodname{}, 32-expert subset\hfill \textcolor{nogray}{\ding{55}}}
]
Raymond was born 6 years before Samantha. So Raymond was born 6 years before Samantha was born. 6 years before Samantha was born was 6 years before 31. So Raymond's son was born 6 ...
\end{tcolorbox}
\begin{tcolorbox}[
  colback=regsoft,colframe=regdark,coltitle=regdark,colbacktitle=regsoft,
  title={Reg.\ MoE, 128-expert subset\hfill \textcolor{nogray}{\ding{55}}}
]
Raymond was born 6 years before Samantha. So he was 6 years older than Samantha. So he was 6 + 31 = 37 years old when his son was born. So the answer is 37.
\end{tcolorbox}
\begin{tcolorbox}[
  colback=emosoft,colframe=emomag,coltitle=emodark,colbacktitle=emosoft,
  title={\methodname{}, 128-expert subset\hfill \textcolor{nogray}{\ding{55}}}
]
Raymond was born 6 years before Samantha. So he was born 6 years before Samantha was 31. So he was born 6 years before 31 = 25 years ago. So the answer is 25 years ago.
\end{tcolorbox}
\end{tcbraster}

%% file: tables/eval_task_details.tex
\begin{table}[t]
\centering
\small
\setlength{\tabcolsep}{4pt}
\begin{tabular}{l c r r l}
\toprule
\textbf{Task} & \textbf{Shots} & \textbf{Train+Val} & \textbf{Test} & \textbf{Primary metric} \\
\midrule
\multicolumn{5}{l}{\textbf{MMLU-Pro} (14 tasks)} \\
\midrule
Math & 5 & 601 & 750 & acc/raw \\
Health & 5 & 388 & 430 & acc/raw \\
Physics & 5 & 580 & 719 & acc/raw \\
Business & 5 & 376 & 413 & acc/raw \\
Biology & 5 & 347 & 370 & acc/raw \\
Chemistry & 5 & 513 & 619 & acc/raw \\
Computer Science & 5 & 224 & 186 & acc/raw \\
Economics & 5 & 398 & 446 & acc/raw \\
Engineering & 5 & 448 & 521 & acc/raw \\
Philosophy & 5 & 260 & 239 & acc/raw \\
Other & 5 & 430 & 494 & acc/raw \\
History & 5 & 213 & 168 & acc/raw \\
Psychology & 5 & 380 & 418 & acc/raw \\
Law & 5 & 501 & 600 & acc/raw \\
\midrule
\multicolumn{5}{l}{\textbf{MMLU} (17 tasks)} \\
\midrule
Biology & 5 & 320 & 182 & acc/raw \\
Business & 5 & 308 & 176 & acc/raw \\
Chemistry & 5 & 211 & 122 & acc/raw \\
Computer Science & 5 & 289 & 165 & acc/raw \\
Culture & 5 & 232 & 134 & acc/raw \\
Economics & 5 & 525 & 298 & acc/raw \\
Engineering & 5 & 103 & 58 & acc/raw \\
Geography & 5 & 140 & 80 & acc/raw \\
Health & 5 & 1\,162 & 659 & acc/raw \\
History & 5 & 658 & 373 & acc/raw \\
Law & 5 & 1\,250 & 707 & acc/raw \\
Math & 5 & 752 & 427 & acc/raw \\
Other & 5 & 825 & 467 & acc/raw \\
Philosophy & 5 & 1\,427 & 808 & acc/raw \\
Physics & 5 & 453 & 257 & acc/raw \\
Politics & 5 & 459 & 260 & acc/raw \\
Psychology & 5 & 823 & 463 & acc/raw \\
\midrule
\multicolumn{5}{l}{\textbf{MC9} (9 tasks)} \\
\midrule
ARC-Easy & 5 & 2\,821 & 2\,376 & acc/raw \\
ARC-Challenge & 5 & 1\,418 & 1\,172 & acc/raw \\
BoolQ & 5 & 9\,427 & 3\,270 & acc/raw \\
HellaSwag & 5 & 39\,905 & 10\,042 & acc/raw \\
CSQA & 5 & 9\,741 & 1\,221 & acc/raw \\
OpenBookQA & 5 & 5\,457 & 500 & acc/raw \\
PIQA & 5 & 16\,113 & 1\,838 & acc/raw \\
SocialIQA & 5 & 33\,410 & 1\,954 & acc/raw \\
WinoGrande & 5 & 40\,398 & 1\,267 & acc/raw \\
\midrule
\multicolumn{5}{l}{\textbf{Gen5} (5 tasks)} \\
\midrule
SQuAD & 5 & 87\,599 & 10\,570 & F1 \\
CoQA & 0 & 108\,647 & 7\,983 & F1 \\
NaturalQS & 5 & 87\,925 & 3\,610 & F1 \\
TriviaQA & 5 & 61\,888 & 7\,993 & F1 \\
DROP & 5 & 77\,409 & 9\,536 & F1 \\
\midrule
\multicolumn{5}{l}{\textbf{GSM8K} (1 task)} \\
\midrule
GSM8K & 8 & 7\,473 & 1\,319 & EM \\
\bottomrule
\end{tabular}
\caption{Evaluation tasks grouped by \textbf{MMLU-Pro} (14 categories), \textbf{MMLU} (17 categories), \textbf{MC9} (9 multiple-choice), \textbf{Gen5} (5 generation), and \textbf{GSM8K}. \textit{Train+Val} reports the size of the dataset used during expert selection and finetuning. }
\label{tab:eval_tasks}
\end{table}